\documentclass[runningheads]{llncs}

\usepackage{accv}

\usepackage{accvabbrv}

\usepackage{graphicx}
\usepackage{booktabs}

\usepackage[accsupp]{axessibility}  

\usepackage{multirow}
\usepackage{threeparttable}    
\usepackage{comment}
\usepackage{wrapfig}

\usepackage{hyperref}

\usepackage{orcidlink}

\begin{document}

\title{Pomo3D: 3D-Aware Portrait Accessorizing\\and More} 


\author{Tzu-Chieh Liu\inst{1} \quad
Chih-Ting Liu\inst{2} \quad
Shao-Yi Chien\inst{1}}


\institute{$^1$National Taiwan University
\qquad $^2$Amazon \\
\email{tzujliu@media.ee.ntu.edu.tw}
}

\maketitle


\begin{abstract}
We propose \textsl{Pomo3D}, a 3D portrait manipulation framework that allows free accessorizing by decomposing and recomposing portraits and accessories. It enables the avatars to attain out-of-distribution (OOD) appearances of simultaneously wearing multiple accessories. Existing methods still struggle to offer such explicit and fine-grained editing; they either fail to generate additional objects on given portraits or cause alterations to portraits (e.g., identity shift) when generating accessories. This restriction presents a noteworthy obstacle as people typically seek to create charming appearances with diverse and fashionable accessories in the virtual universe. Our approach provides an effective solution to this less-addressed issue. We further introduce a “Scribble2Accessories” module, enabling \textsl{Pomo3D} to create 3D accessories from user-drawn accessory scribble maps. Moreover, we design a bias-conscious mapper to mitigate biased associations present in real-world datasets. In addition to object-level manipulation above, \textsl{Pomo3D} also offers extensive editing options on portraits, including global or local editing of geometry and texture and avatar stylization, elevating 3D editing of neural portraits to a more comprehensive level. Project page: \url{https://tzuj6.github.io/Pomo3D}.
  
\end{abstract}

\section{Introduction}
\label{sec:intro}

Recently, with the swift development of virtual reality (VR) and augmented reality (AR), portrait generation has found various promising applications, such as avatar-based telepresence or teleconference, providing users with immersive experiences. Thus, producing 3D portrait images that are highly realistic and editable has been a surge of interest in recent years.

2D GANs, such as the prevalent StyleGAN-based approaches~\cite{abdal2021styleflow,chen2020deepfacedrawing,Deng_2020_CVPR,shen2020interfacegan,Tewari_2020_CVPR,Zhu_2020_CVPR}, can achieve impressive portrait generation and manipulation. However, they inherently disregard fundamental principles of projective geometry, thus leading to incoherent editing when the viewpoint shifts. To address this issue, several prior work~\cite{Cai_2022_CVPR,chan2021pi,gu2021stylenerf,niemeyer2021giraffe,or2022stylesdf,schwarz2020graf} have employed 3D GANs to yield view-consistent results based on the 3D-structure-aware inductive bias introduced by the neural rendering pipeline. Additionally, given the promising results of diffusion models~\cite{sohl2015deep} on 2D images, some studies~\cite{wang2022rodin,gu2024control3diff,lan2023gaussian3diff,kirschstein2024diffusionavatars,karnewar2023holofusion} integrate denoising processes with 3D representations, enabling diffusion models to grasp an understanding of the underlying 3D structure. 

Nevertheless, no tailored work has been done on 3D portrait accessorizing that solely uses easily accessible 2D datasets for training, and the granularity of manipulation in existing 3D generative models is still insufficient for accessory wearing, which significantly limits the variety of personalized avatars' appearances. Generally, they model both portraits and relatively uncommon accessories (e.g., earrings) in a single neural scene representation, leading to two potential outcomes. \textbf{1}) Due to severe data imbalance between portraits and accessories, models may fail to effectively generate these less common objects. \textbf{2}) It may result in significant object entanglement. Thus, altering accessories easily leads to changes in the wearer.

\begin{figure}[tb]
  \includegraphics[width=\textwidth]{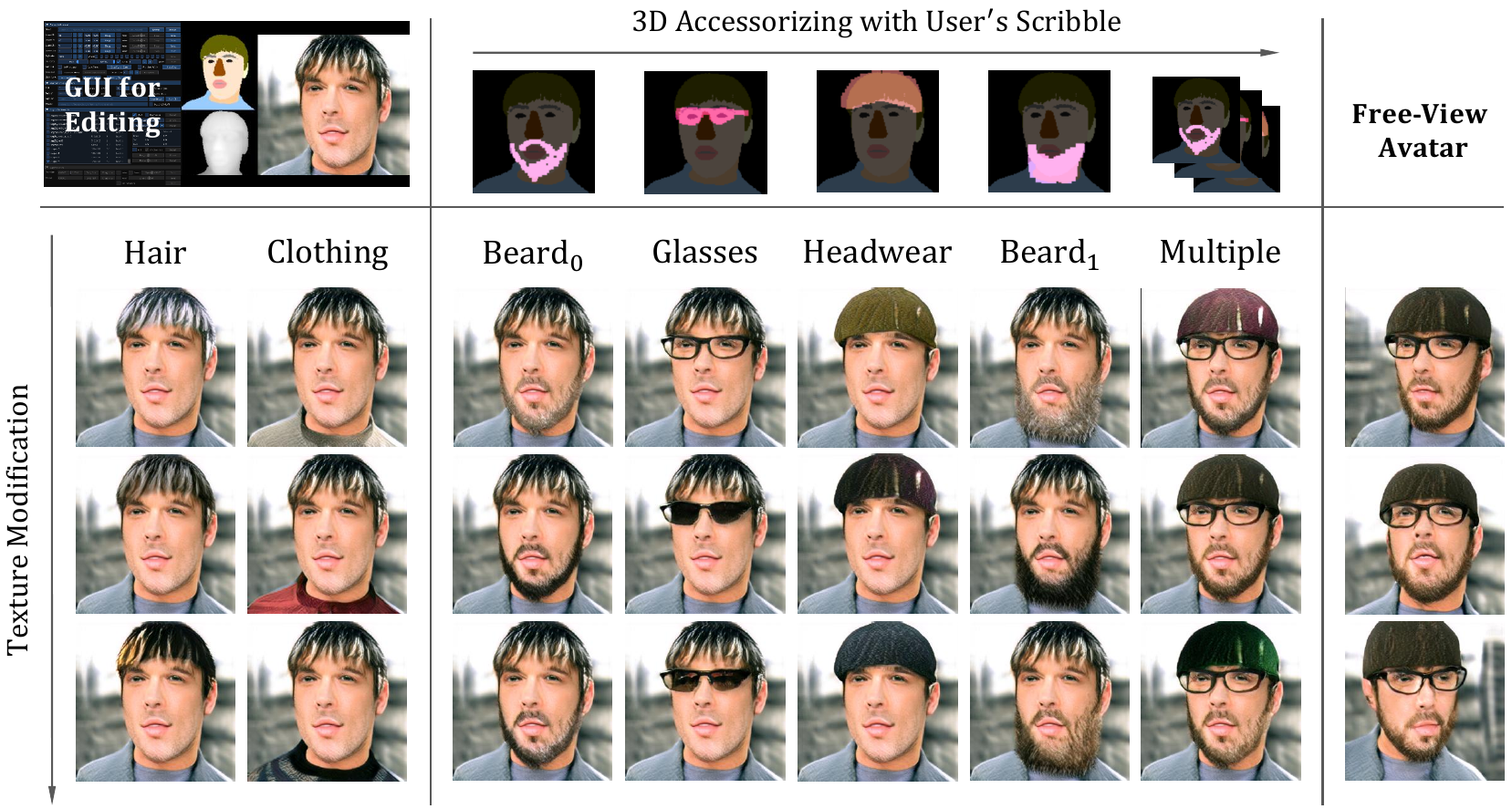}
  \caption{Through user-drawn shapes of scribbles (top row) and diverse texture selections, \textsl{Pomo3D} can generate personalized accessories on specified avatars (top left in GUI). These scribble maps can be directly drawn within our provided GUI, and multiple scribbles can be stacked together to achieve multiple accessories worn concurrently. All the results are multi-view consistent and operate at an interactive frame rate. 
  }
  \label{fig:teaser}
\end{figure}

To address vital necessities, we propose the \textsl{Pomo3D} framework enabling not only high-degree freedom in accessory editing but also extensive portrait editing options, encompassing geometry modification, global or local texture editing, and avatar stylization. We establish two distinct scene representations for modeling geometry; one is for portraits, and the other is tailored for less common accessories. A novelly designed feature adapter is utilized to connect these two representations. Following this, we obtain projected feature maps and segmentation masks from two geometry representations through volume rendering. These projected geometry features of portraits and accessories are fused and then texture-rendered into high-resolution RGB images, and the segmentation masks serve as semantic-aware constraints, guiding the feature fusion and texture rendering. Through such a design, we can inject explicit and precise semantic control into the architecture, introduce an inductive bias that disentangles geometry and texture, and decompose portraits and accessories simultaneously. To the best of our knowledge, \textsl{Pomo3D} offers the highest level of editability for decorative objects. As illustrated in \cref{fig:teaser}, control over accessories and beards includes geometry and texture modification, generation from the user's unrestricted scribbles, and concurrent wearing of multiple accessories. 

Due to the lack of dedicated accessory datasets and annotations, we first extract accessories from existing datasets and organize them into the \textbf{PAC}-Mask (\textbf{P}ersonal \textbf{AC}cessories) dataset to train \textsl{Pomo3D}. This dataset can support extensive experiments on portrait accessorizing with our approach or future studies. For better modeling, we also consider the strong bias inherent in the real-world dataset and propose a bias-conscious mapper, which more faithfully models pose-dependent biases and identity-correlated biases when putting on accessories. Furthermore, “Scribble2Accessories” is introduced to allow \textsl{Pomo3D} to accessorize from scribble maps. These scribbles can either be detailed strokes drawn within the GUI tool or rough sketches made by the user's fingertip on a touchscreen. Since accessories and portraits are modeled separately, re-combining them can produce images beyond the representation of the dataset, such as men having the option to wear various earrings. When combining portraits and accessories, interactions between them (e.g., deformation) are also considered, as the geometry of accessories is constructed based on the portrait's geometry.

In essence, our primary technical contributions are as follows:
\begin{itemize}
\item[$\bullet$] We propose the first 3D portrait manipulation framework allowing for independent control at the object level (e.g., accessories, beard) and extensive editing options, including geometry modification, intra- or inter-domain stylization, global or local texture editing, etc.
 
\item[$\bullet$] We devise a “Scribble2Accessories” module, which can directly create 3D accessories based on the user’s unconstrained accessory scribble.

\item[$\bullet$] We leverage a bias-conscious mapper to disconnect biased associations in real-world datasets. This facilitates the segregation of accessories from portraits and generalization beyond training distribution, attaining \\
out-of-distribution (OOD) accessorizing.

\item[$\bullet$] We create a PAC-Mask dataset curated from existing datasets to support extensive portrait accessorizing experiments with our approach or future studies, eliminating the need for in-house or multi-view data of accessories.
\end{itemize}

\section{Related Work}

\subsection{Neural Scene Representation}
A scene representation incorporating a neural network to approximate surface or volumetric representation functions is commonly referred to as \textit{neural scene representation} within the field of neural rendering~\cite{tewari2022advances,liao2024advances}. These scenes can be represented either implicitly or explicitly. In the case of volumetric representation, the quantities of interest (occupancies, colors, etc.) of the volume can be either represented by voxel grids~\cite{sitzmann2019deepvoxels, peng2020convolutional, kanazawa2018learning} or defined by the weights of a neural network~\cite{mildenhall2021nerf, gafni2021dynamic, park2021nerfies}. A hybrid approach combining explicit and implicit representations is another way to balance memory usage and inference speed~\cite{xu20223d, chan2022efficient}. In this work, we opt for the tri-plane hybrid scene representation and build upon their work~\cite{chan2022efficient}, which has faster inference speed and scales efficiently with resolution.

\subsection{3D-Aware Generative Models}
Equipped with the neural scene representation and a differentiable rendering algorithm, 3D-aware GANs can produce multi-view consistent images~\cite{schwarz2020graf, chan2021pi, Tewari_2022_CVPR}. Several approaches \cite{gu2021stylenerf, or2022stylesdf, zhou2021cips, chan2022efficient, xue2022giraffe} adopt a two-stage rendering process, which leverages convolution neural networks (CNNs) to increase the resolution of the image or neural rendering features, to generate 3D-aware images at higher resolution efficiently. In addition, GIRAFFE~\cite{niemeyer2021giraffe}, gCoRF~\cite{br2022gcorf}, and CNeRF~\cite{ma2023semantic} employ a compositional neural radiance field that supports object-level controls. However, GIRAFFE falls short in explicitly separating objects of interest, while gCoRF and CNeRF neglect handling uncommon objects (e.g., earrings, headwear, etc.).

\subsection{Controllable Portrait Synthesis}
With the advent of generative models, researchers are also intrigued by the capacity for portrait manipulation. However, in style-based generators, the meanings of latent codes still need to be clarified, making them difficult to manipulate. SemanticStyleGAN~\cite{shi2022semanticstylegan} utilizes compositional synthesis to encourage spatial disentanglement, attaining photo-realistic images and finer-grained control. Nevertheless, akin to other 2D mask-based editing methods~\cite{lee2020maskgan,Zhu_2020_CVPR,zhu2021barbershop}, it lacks 3D knowledge, yielding incoherent editing when the viewpoint shifts. SofGAN~\cite{chen2022sofgan} produces perspective-consistent semantic maps with a semantic occupancy field, achieving independent control over portrait shape and texture. Nonetheless, it only ensures semantic-level 3D consistency and requires additional labeled multi-view data. By learning the joint distribution of images and semantics, FeNeRF \cite{sun2022fenerf} utilizes the semantic branch as an intermediary to edit color images via GAN inversion and achieves multi-view consistent and regional editing. However, owing to the extensive computational demand, the image quality is sub-optimal. IDE-3D~\cite{sun2022ide} utilizes a pair of tri-planes to disentangle shape and appearance attributes, resulting in portrait editing capabilities that are both flexible and 3D consistent. Nonetheless, it restricts editing only on a fixed view and may 
produce artifacts at the semantic boundary when locally editing the texture. pix2pix3D~\cite{deng20233d} maps semantic graphs onto the generator's latent space, facilitating controllable graph-to-image generation. However, its geometry and texture spaces are highly entangled, which means that altering the semantic maps easily leads to variations in texture.

While there have been promising advancements in 3D portrait manipulation~\cite{jiang2022nerffaceediting,sun2022next3d,li20243d,kim2024diffusion,chen2024hyperstyle3d,li2023instructpix2nerf,hyung2023local,zhou2023mate3d,cha2024pegasus,li2023preim3d,gao2023sketchfacenerf,sun2023make}, existing approaches still cannot explicitly incorporate a diverse range of accessories into a given 3D portrait, which significantly restricts the diversity of virtual avatars. By contrast, \textsl{Pomo3D} tackles the issues mentioned above and offers broader editing options that encompass not only shape and texture control at both global and local levels but also operations at the object level.

\begin{figure}[tb]
  \includegraphics[width=\textwidth]{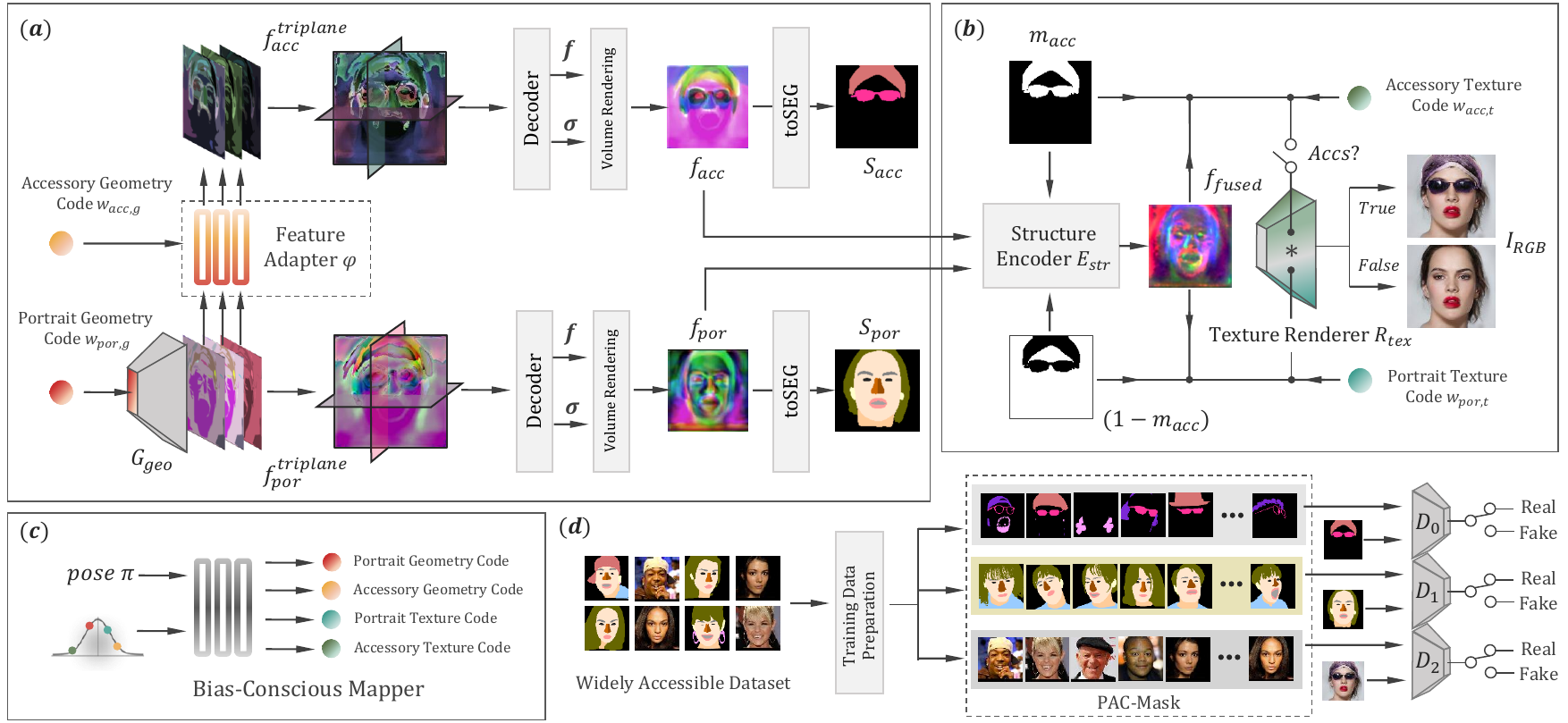}
  \caption{\textbf{Overview.}
  (a) Generation of dual geometry tri-planes: we construct two tri-planes for the geometry modeling of portraits and accessories. We then obtain the projected feature maps and corresponding semantic maps via volume rendering.
  (b) Structure-guided texture renderer: next, the structure encoder and texture renderer fuse the two projected feature maps and yield the output image. The variable $\mathnormal{Accs}$ indicates whether the accessory is worn on the portrait, and thus there are two possible outcomes.
  (c) Bias-conscious mapper: considering the biases in existing datasets, a bias-conscious mapper is proposed to map Gaussian noise into four latent codes for corresponding attributes.
  (d) Data preparation and training scheme: PAC-Mask consists of three data groups: accessory semantic maps, portrait semantic maps, and overall RGB images. During training, we use three different discriminators along with these three data groups to conduct adversarial learning on the three branches of the network.
  }
  \label{fig:F3}
\end{figure}

\section{Methodology}
\subsubsection{Overview}
As shown in~\cref{fig:F3}, we first utilize a StyleGAN2 generator $\mathnormal{G}_{geo}$ to construct the geometry tri-plane of portraits $\mathnormal{f}_{por}^{triplane}$ with the portrait geometry code $\mathnormal{w}_{por,g}$, and a feature adapter maps the features from the portrait space to the accessory space with the accessory geometry code $\mathnormal{w}_{acc,g}$, thereby establishing another geometry tri-plane for the accessory $\mathnormal{f}_{acc}^{triplane}$. Two projected feature maps ($\mathnormal{f}_{por}, \mathnormal{f}_{acc}$) and corresponding 
semantic maps ($\mathnormal{S}_{por}, \mathnormal{S}_{acc}$) can be obtained via volume rendering and a per-pixel classifier $\mathnormal{F}_{toSEG}$. Next, the structure encoder $\mathnormal{E}_{str}$ fuses two geometry feature maps ($\mathnormal{f}_{por}, \mathnormal{f}_{acc}$), forming a structural prior for the texture renderer $\mathnormal{R}_{tex}$. Lastly, the object-aware texture renderer $\mathnormal{R}_{tex}$, modulated by two latent codes controlling the texture ($\mathnormal{w}_{por,t}, \mathnormal{w}_{acc,t}$), produces the final output $\mathnormal{I}_{RGB}$.

\begin{figure}[tb]
  \centering
  \includegraphics[width=0.5\linewidth]{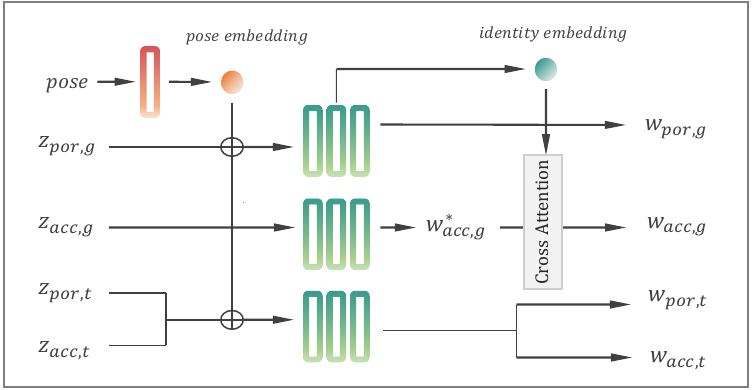}
  \caption{Bias-Conscious Mapper. Our mapping network generates style codes that are aware of both pose and identity, through pose conditioning and identity conditioning. $\mathcal{W}_{acc,g}^{*}$ and $\mathcal{W}_{acc,g}$ are identity-uncorrelated and identity-correlated space, respectively.}
  \label{fig:F4}
\end{figure}

\subsection{Bias-Conscious Mapper}
\label{sec:bias mapper}
Following StyleGAN2 \cite{karras2020analyzing}, to handle the non-linearity of data distribution, a noise vector \textbf{z} from the spherical Gaussian space $\mathcal{Z}$ is first to be mapped into an intermediate style space $\mathcal{W}$ with Multi-Layer
Perceptron (MLP) layers and be extended into a $\mathcal{W}^{+}$ space. However, unlike most methods, we attempt to decompose the $\mathcal{W}^{+}$ space into four subspaces: portrait geometry, accessory geometry, portrait texture, and accessory texture. The factorized $\mathcal{W}^{+}$ space can be formalized as:
\begin{equation}
\mathcal{W}^{+} = \mathcal{W}_{geometry} \times \mathcal{W}_{texture} = \mathcal{W}_{por,g} \times \mathcal{W}_{acc,g} \times \mathcal{W}_{por,t} \times \mathcal{W}_{acc,t} 
\end{equation} 
where the subscripts \textit{por}, \textit{acc}, \textit{g} and \textit{t}, refer to \textit{portrait}, \textit{accessory}, \textit{geometry} and \textit{texture}, respectively. As depicted in~\cref{fig:F3}(c), due to the decomposition of our latent space into four distinct subspaces, each managed by a unique code, we greatly increase flexibility in manipulating the avatar.

Nevertheless, most real-world datasets contain biases. Unsophisticated handling of these biases leads to unfavorable results and entangled attributes. In this work, we mainly address two significant biases. Firstly, for the task that learns 3D consistent views from a set of arbitrary single-view images, the model is prone to capture spurious correlations between poses and appearances (e.g., expression, gaze direction) due to ambiguities between viewpoints. Secondly, for the task of producing a diverse combination of accessories and wearers, due to the model's limited observation of real-world data, it tends to incorrectly associate accessories with portraits of specific attributes. For instance, the proportion of males wearing earrings is much lower than that of females, demonstrating an accessory-gender association. Faithfully modeling these attribute correlations during training and decoupling these intertwined attributes during inference is necessary to achieve both multi-view consistent and diverse accessories wearing.

To this end, we propose a bias-conscious mapper handling pose-appearance associations and accessory-portrait associations as in~\cref{fig:F4}. Following~\cite{chan2022efficient}, we provide the network with knowledge of camera poses. Furthermore, we employ a cross-attention module to make the generation of accessories identity-aware, thereby better reconstructing the training data distribution. During training, however, instead of fully sampling the accessory geometry code from the identity-correlated space $\mathcal{W}_{acc,g}$, we randomly sample from both spaces $\mathcal{W}_{acc,g}^{*}$ and $\mathcal{W}_{acc,g}$ with a probability of \textit{p}. This is clarified in the following experiments in~\cref{sec:Identity Conditioning}. During inference, to decouple pose- and identity-related attributes, a fixed camera pose is conditioned, and we entirely sample the accessory geometry code from the identity-uncorrelated space $\mathcal{W}_{acc,g}^{*}$.

\subsection{Generation of Dual Geometry Tri-planes}
As depicted in~\cref{fig:F3}(a), we first learn a StyleGAN-based generator $\mathnormal{G}_{geo}$ to generate a portrait geometry tri-plane $\mathnormal{f}_{por}^{triplane} \in \mathbb{R}^{3 \times 32 \times 256 \times 256}$ with a portrait geometry code $\mathnormal{w}_{por,g}$, following the setting in EG3D~\cite{chan2022efficient}. In general, given the structure of a face, people can easily imagine where accessories can be placed (e.g., earlobes, top of the head). Hence, our key insight is that, for an already established portrait geometry representation, a lightweight model can be introduced to infer the accessory geometry representation based on the knowledge of the portrait's structure. We implement this model with a compact feature adapter $\varphi$, which consists of three branches $(\varphi_{a}, \varphi_{b}, \varphi_{c})$, where $\varphi_{i}: \mathnormal{f}_{por}^{plane} \in \mathbb{R}^{32 \times 256 \times 256} \mapsto \mathnormal{f}_{acc}^{plane} \in \mathbb{R}^{32 \times 256 \times 256}$, mapping the features of each 32-channel plane from portrait to the accessory space, conditioned on the accessory geometry code $\mathnormal{w}_{acc,g}$. As such, dual geometry tri-planes $(\mathnormal{f}_{por}^{triplane}, \mathnormal{f}_{acc}^{triplane})$ can be constructed. The features $\mathnormal{f}$ and density $\sigma$ of a 3D point can be queried by projecting the point onto three orthogonal planes, aggregating the features from three planes, and processing by a decoder. Then we can project the 3D feature volume onto 2D feature images $\mathnormal{f}_{por}, \mathnormal{f}_{acc} \in \mathbb{R}^{64 \times 128 \times 128}$ via volume rendering:

\begin{equation}
\mathnormal{f}_{por} = \xi \left( \mathnormal{G}_{geo} \left( \mathnormal{w}_{por,g} \right), \pi \right)
\quad
\mathnormal{f}_{acc} = \xi \left( \varphi \left( \mathnormal{G}_{geo} \left( \mathnormal{w}_{por,g} \right), \mathnormal{w}_{acc,g} \right), \pi \right)
\end{equation}

\begin{figure}[tb]
  \centering
  \includegraphics[width=0.8\linewidth]{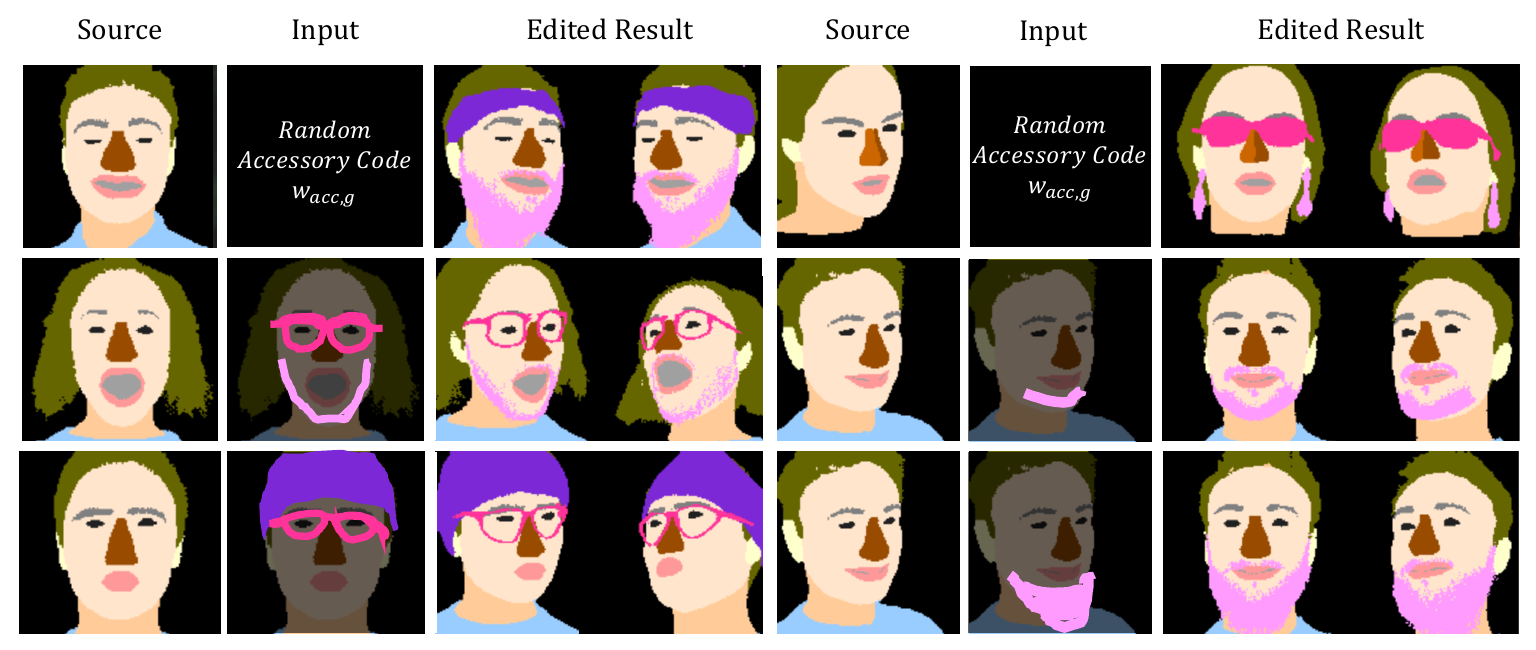}
  \caption{
  Accessories and beards can be generated either by a random accessory geometry code (first row) or by the user's scribble map (bottom two rows). They can be created from any viewpoint, not limited to the frontal view. The two examples in the bottom right corner demonstrate that different stroke widths lead to different types of beards.
  }
  \label{fig:F5}
\end{figure}

Here, $\xi$ denotes the neural volume rendering process, given a camera pose $\pi$. We further extract the semantic maps $\mathnormal{S}_{por} \in \mathbb{R}^{20 \times 128 \times 128}$ and $\mathnormal{S}_{acc} \in \mathbb{R}^{5 \times 128 \times 128}$, by predicting the semantics of each pixel on the feature images via a MLP classifier $\mathnormal{F}_{toSEG}$, which describes the probabilistic distribution over \textit{N} semantic classes (such as background, skin, cloth, etc.). As the structure of the feature images is so well-defined, the classifier $\mathnormal{F}_{toSEG}$ can be designed to be extremely lightweight. As shown in~\cref{fig:F5}, the accessories inferred from the portrait geometry will implicitly and correctly align with the portraits without imposing additional constraints.

\subsection{Structure-Guided Texture Renderer}
This section consists of two parts: the structure encoder $\mathnormal{E}_{str}$ and the texture renderer $\mathnormal{R}_{tex}$. The structure encoder is proposed to capture the essential structural information from the two feature images $(\mathnormal{f}_{por}, \mathnormal{f}_{acc})$, to form the structural prior for texture generation. It utilizes a pair of binary masks, $\mathnormal{m}_{acc}$ and $(1-\mathnormal{m}_{acc})$, that indicate the accessory and non-accessory regions, along with the two feature images, to output a fused feature map $\mathnormal{f}_{fused} = \mathnormal{E}_{str} \left( \mathnormal{f}_{por}, \mathnormal{f}_{acc}, \mathnormal{m}_{acc} \right) \in \mathbb{R}^{32 \times 128 \times 128}$. This makes it another “adapter” between the geometry and texture space, yielding higher image quality while maintaining view consistency.

For the texture renderer $\mathnormal{R}_{tex}$, the fused feature map $\mathnormal{f}_{fused}$ serves as a structural prior, which can be regarded as an extra conditional variable, guiding the generation of fine-grained textures. Assuming that natural images belong to a joint distribution $\mathcal{F}(\mathcal{G},\mathcal{T})$ of geometry $\mathcal{G}$ and texture $\mathcal{T}$, typically, a image generator $\mathnormal{G}$ learns: $\mathnormal{p}_{z} \stackrel{\mathnormal{G}}{\longrightarrow} \mathcal{F}(\mathcal{G},\mathcal{T})$, while the texture renderer $\mathnormal{R}_{tex}$ learns: $\mathnormal{p}_{z_{t}} \stackrel{\mathnormal{R}_{tex}}{\longrightarrow} \mathcal{F}(\mathcal{T}|\mathcal{G})$, given a structural prior as the known geometry. 

To construct the texture renderer, built on~\cite{chen2022sofgan}'s insight, we first alter the StyleGAN2 block to function as a region-aware texture rendering block and use these blocks to upsample and modulate with two texture styles $(\mathnormal{w}_{por,t}, \mathnormal{w}_{acc,t})$. Specifically, we adopt a compositional synthesis procedure within each texture rendering block $\mathnormal{R}^{i}$. Features in each block are modulated by two styles and then re-combined based on binary masks $\mathnormal{m}_{acc}$ and $(1 - \mathnormal{m}_{acc})$:

\begin{equation}
\mathnormal{f}^{n} = \mathnormal{m}_{acc} \odot \mathnormal{R}^{n}_{w_{acc,t}} \left( \mathnormal{f}^{n-1} \right) + \left( 1 - \mathnormal{m}_{acc} \right) \odot \mathnormal{R}^{n}_{w_{por,t}} \left( \mathnormal{f}^{n-1} \right)
\end{equation}

\begin{figure}[tb]
  \centering
  \includegraphics[width=0.95\linewidth]{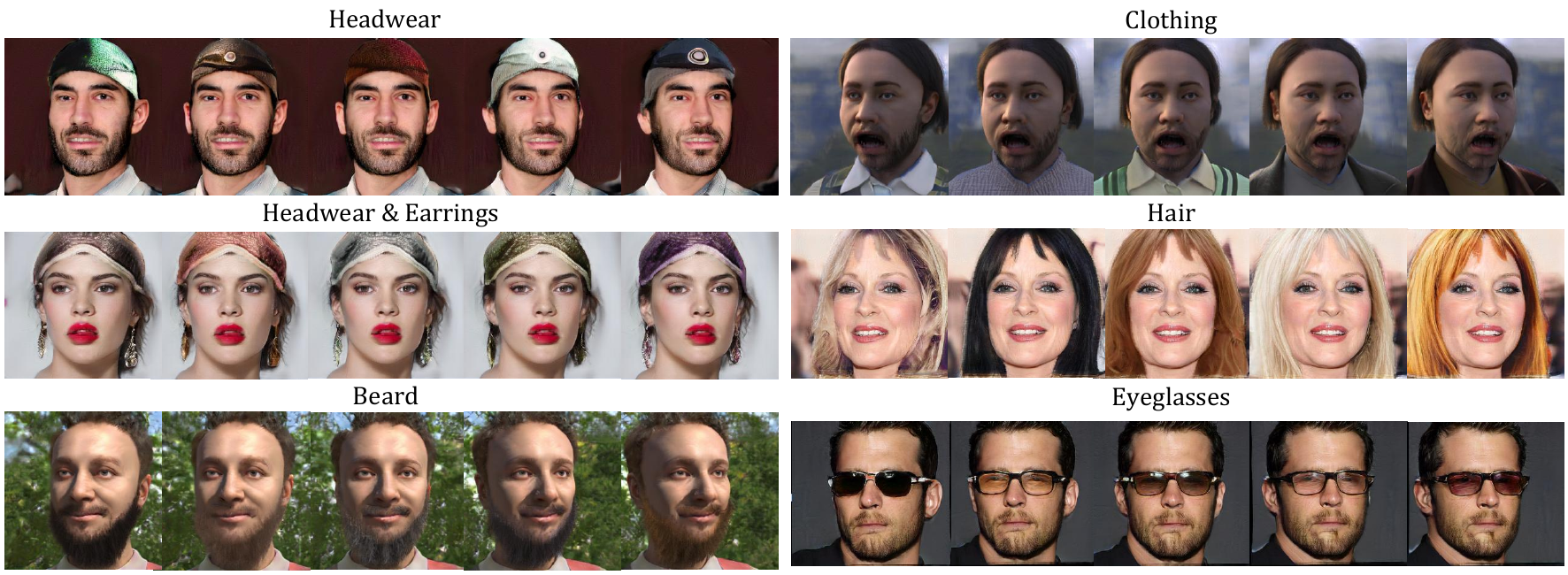}
  \caption{
  The high degree of regional disentanglement facilitates fine-grained texture adjustments for each object, allowing for extensive customization of avatar accessories.
  }
  \label{fig:F23}
\end{figure}

Here, $\odot$ denotes element-wise multiplication. $\mathnormal{f}^{n}$ and $\mathnormal{f}^{n-1}$ are the output features and input features of the $\mathnormal{n}^{th}$ texture rendering block, respectively. $\mathnormal{R}^{n}_{w}$ represents the $\mathnormal{n}^{th}$ rendering block whose kernel weights are modulated by the texture style $\mathnormal{w}$. For the first rendering block $\mathnormal{R}^{1}$, input features $\mathnormal{f}^{0}$ corresponds to the fused feature map $\mathnormal{f}_{fused}$. Employing such a regional linear blending procedure substantially enhances regional-level disentanglement, implicitly enabling the model to integrate different texture styles at the object level seamlessly, as demonstrated in~\cref{fig:F23}. Also, under the guidance of the fused feature map $\mathnormal{f}_{fused}$ as a structural prior, the texture generation guarantees strong multi-view consistency. We further introduce spatially-adaptive normalization~\cite{park2019semantic} into our rendering block to impose semantic-aware constraints on generated features with semantic maps $(\mathnormal{S}_{por}, \mathnormal{S}_{acc})$. For situations where accessories are not required (i.e., $\mathnormal{Accs} == False$), the binary mask $\mathnormal{m}_{acc}$ will be an all-zeros mask, indicating the fused features $\mathnormal{f}_{fused}$ will be derived solely from portrait features $\mathnormal{f}_{por}$, and only the portrait texture code $\mathnormal{w}_{por,t}$ will be used during texture rendering. In practice, we may extend the binary mask $\mathnormal{m}_{acc}$ to regions beyond accessories (e.g., clothing, hair). Please refer to the supplement.

\subsection{Data Preparation and Training Scheme}
\subsubsection{Data Preparation}
We curate PAC-Mask from the widely accessible datasets, CelebAMask-HQ~\cite{lee2020maskgan}, FFHQ~\cite{karras2019style} and FaceSynthetics~\cite{wood2021fake}. The main preprocessing involves emphasizing the accessory region of segmaps (segmentation maps), splitting the nose semantics into two parts for clearer geometry, and conducting pose detection. Next, we categorize the segmaps into two groups based on the presence of accessories and extract the accessory semantics. Consequently, the PAC-Mask dataset comprises three non-overlapping groups: accessory segmaps, unadorned portrait segmaps, and RGB images. Further details and statistics are included in the supplementary materials.

\subsubsection{Training Scheme}
As illustrated in~\cref{fig:F3}(d), we novelly employ a three-pronged strategy to train \textsl{Pomo3D}, specifically by using three discriminators to supervise three corresponding branches separately. The three real data groups from PAC-Mask and the generated results by three branches, $\mathnormal{S}_{acc}$, $\mathnormal{S}_{por}$ and $\mathnormal{I}_{RGB}$, are used for adversarial learning independently. The architecture of the three discriminators is identical to those of EG3D~\cite{chan2022efficient}, with the only difference being the input channel (for example, 5 for the discriminator $\mathnormal{D}_{0}$ that discriminates accessory segmaps). Note that two branches for segmaps $\mathnormal{S}_{acc}$ and $\mathnormal{S}_{por}$ inject two shorter gradient backward paths, which makes the training more stable. This strategy explicitly aligns the distribution of each branch's output with that of the corresponding data group.

\begin{figure}[tb]
  \centering
  \includegraphics[width=0.7\linewidth]{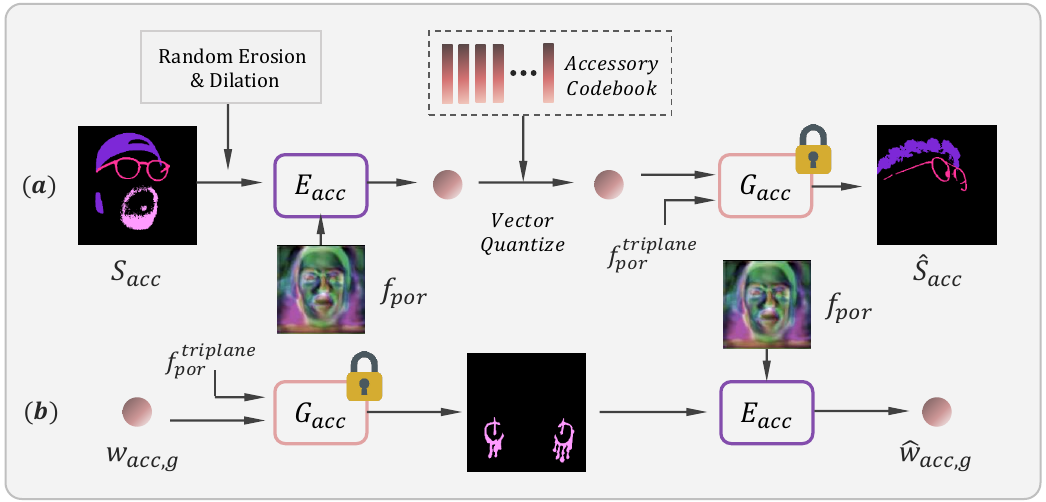}
  \caption{
  Training of Scribble2Accessories. We train the encoder $\mathnormal{E}_{acc}$ and the accessory codebook with two types of cycle consistency while fixing the pre-trained generator.
  }
  \label{fig:F6}
\end{figure} 


\subsection{Scribble2Accessories Inversion}
What sets us apart from most existing approaches is our capability to conduct GAN inversion on specified objects while keeping the remaining part unchanged, thanks to the design of the dissociated branch tailored for accessories. We introduce a Scribble2Accessories module to synthesize photo-realistic 3D accessory assets from inaccurate hand-drawn scribbles. In practice, we first design a 2D encoder $\mathnormal{E}_{acc}$ that maps scribbles to the accessory geometry code $\mathnormal{w}_{acc,g}$, providing the encoder with projected portrait features $\mathnormal{f}_{por}$ as the pose and structural information of the wearer, namely, $\mathnormal{w}_{acc,g} = \mathnormal{E}_{acc} \left( \mathnormal{S}_{acc}, \mathnormal{f}_{por}  \right)$. During training, $\mathnormal{S}_{acc}$ is an actual accessory segmap (either sampled from the training data group or generated by the pre-trained generator). During inference, $\mathnormal{S}_{acc}$ will be a hand-drawn accessory segmap. To mitigate the domain gap between the two, we implement $\mathnormal{E}_{acc}$ as a VQ-based (vector-quantized)~\cite{esser2021taming,van2017neural} encoder and build a discrete codebook $\mathcal{A} = \left\{ \mathnormal{w}_{k} \right\}_{k=1}^{K} \subset \mathbb{R}^{n_{w}}$ of learned accessory representation. Thus, the mapped latent $\mathnormal{w}_{acc,g}$ is then quantized onto its closest accessory codebook entry $\mathnormal{w}_{k}$. This ensures that all predicted accessory geometry codes during inference lie within the distribution of actual accessory geometry codes.

Let $\mathnormal{G}_{acc}$ denote pre-trained modules in the accessory branch (upper part of~\cref{fig:F3}(a)), including the feature adapter $\varphi$, the neural volume rendering process, and its per-pixel classifier $\mathnormal{F}_{toSEG}$. A reconstructed accessory segmap $\hat{\mathnormal{S}}_{acc} = \mathnormal{G}_{acc} (\mathnormal{w}_{k}, \mathnormal{f}_{por}^{triplane})$ can then be produced given the quantized accessory geometry code $\mathnormal{w}_{k}$ and the portrait tri-plane features $\mathnormal{f}_{por}^{triplane}$. To train the encoder $\mathnormal{E}_{acc}$ and the accessory codebook $\mathcal{A}$, as depicted in~\cref{fig:F6}, we use two types of cycle consistency, $(\mathnormal{a})$ and $(\mathnormal{b})$, while fixing the generator $\mathnormal{G}_{acc}$. In $(\mathnormal{a})$, we sample accessory segmaps from the data and randomly apply the erosion or dilation in cv2~\cite{opencv_library} to handle hand-drawn segmaps. In $(\mathnormal{b})$, a random accessory geometry code $\mathnormal{w}_{acc,g}$ is sampled and used to calculate loss with the reconstructed code $\hat{\mathnormal{w}}_{acc,g}$. Let $\mathcal{L}_{recon}$ 
denote a reconstruction loss to minimize segmap discrepancy, and $\mathcal{L}_{latent}$ denote a smooth L1 latent loss. The overall learning objective is:

\vspace{-9pt}
\begin{equation}
\theta_{E_{acc}}^{*} = \mathop{\arg\min}\limits_{\theta_{E_{acc}}}  \sum_{n}^{N} \mathcal{L}_{recon} \left( \mathnormal{S}_{acc}^{n}, \hat{\mathnormal{S}}_{acc}^{n} \right) + \alpha \cdot \mathcal{L}_{com} \left( \mathnormal{E}_{acc}, \mathcal{A} \right) + \beta \cdot \mathcal{L}_{latent} \left( \mathnormal{w}_{acc,g}^{n}, \hat{\mathnormal{w}}_{acc,g}^{n} \right)
\end{equation}

Here $\mathcal{L}_{com}$ is the so-called “commitment loss” as in~\cite{esser2021taming,van2017neural}. \textit{N} is the total number of observed samples within a batch. With the encoder $\mathnormal{E}_{acc}$ and the pre-trained generator $\mathnormal{G}_{acc}$, we can convert a user's 2D scribble into free-view accessories, as illustrated in the bottom two rows of \cref{fig:F5}.

\begin{figure}[tb]
  \centering
  \includegraphics[width=\linewidth]{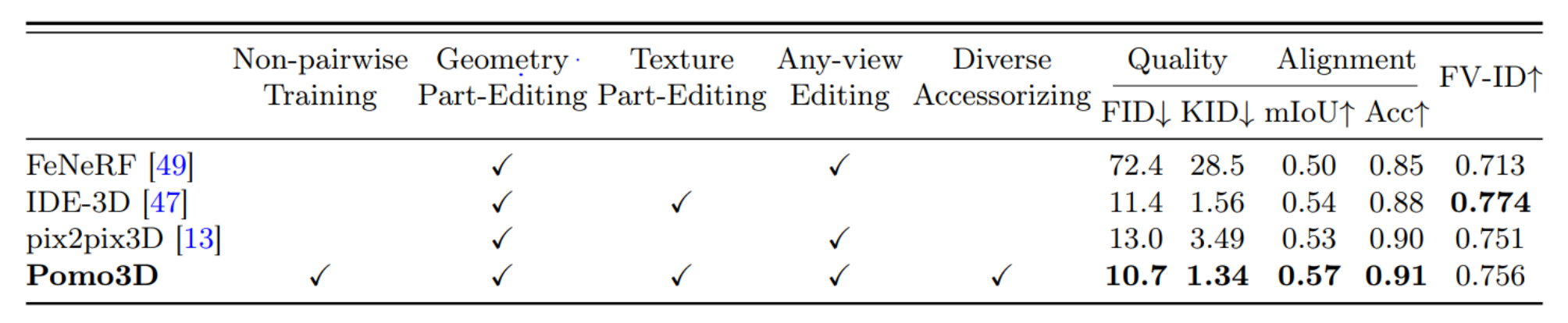}
  \caption{
  Comparison with SOTA masked-based 3D portrait editing methods. \textit{Non-pairwise Training} refers to the fact that RGB-segmap pairwise training is not required, providing greater training data flexibility. \textit{Any-view Editing} represents the capability to edit from any viewpoint. \textit{Diverse Accessorizing} includes various types of accessory wearing, fine-grained control over accessories, etc.
  }
  \label{fig:table}
\end{figure}

\section{Experiments}

\subsection{Evaluation Metrics}
To evaluate the image quality of synthesized images, we use FID (Fréchet Inception Distance)~\cite{heusel2017gans} and KID (Kernel Inception Distance)~\cite{binkowski2018demystifying} to measure the distribution distance between generated images and real images. To evaluate the alignment between output RGB images and output segmaps, we first predict the semantics of RGB images with an off-the-shelf network~\cite{yu2018bisenet} and then calculate mIoU (mean Intersection-over-Union) and Acc (pixel accuracy) between output segmaps and the predicted semantics following~\cite{park2019semantic}. For view consistency evaluation, we assess the preservation of facial identity across different viewpoints with FV-ID (free-view facial identity consistency), which calculates the mean cosine similarity of Arcface~\cite{deng2019arcface} features.

\subsection{Comparison}

\subsubsection{Quantitative evaluations}
We compare \textsl{Pomo3D} with other SOTA mask-based 3D portrait editing models using their official codes. As shown in~\cref{fig:table}, in addition to providing more flexible training data setups, our proposed method outperforms FeNeRF\cite{sun2022fenerf}, IDE-3D\cite{sun2022ide}, and pix2pix3D\cite{deng20233d} in editing flexibility, image quality, and alignment while being competitive with IDE-3D and pix2pix3D in view consistency.

\subsubsection{Qualitative evaluations}
\begin{figure}[tb]
  \centering
  \includegraphics[width=\linewidth]{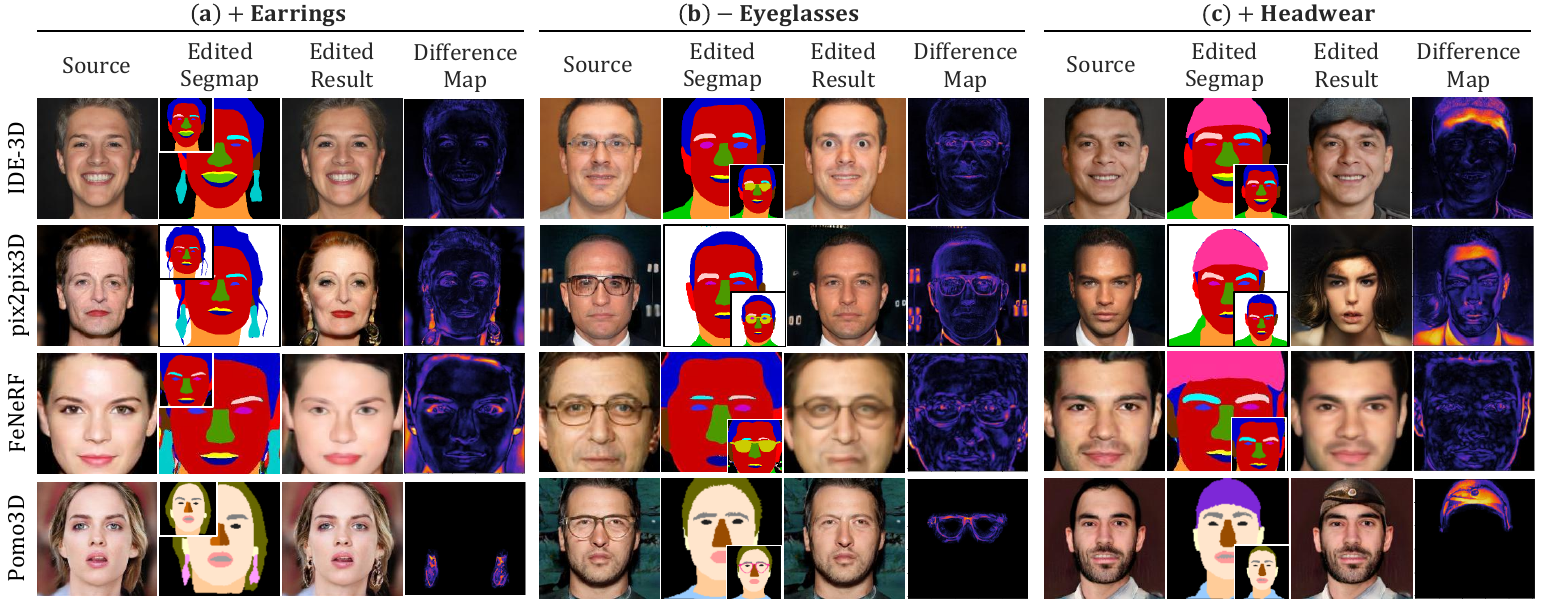}
  \caption{
  A visual comparison of accessory editing capability and corresponding impact regions. For each method, we define three accessory editing tasks: (a) wearing earrings, (b) removing eyeglasses, and (c) donning a hat. In the column of \textit{Edited Segmap}, the smaller and the larger segmap show before and after editing segmaps. \textit{Difference Map} presents pixel differences between the source and the edited result.
  }
  \label{fig:F8}
\end{figure}

In~\cref{fig:F8}, we compare the editing capability of accessories and the impact region of editing with SOTAs. It is observed that IDE-3D still struggles with tasks beyond eyeglass editing; for instance, in task(a), their result more closely resembles braids rather than earrings, and in task(c), it mistakenly generates hair instead of headwear. pix2pix3D can edit earrings; however, the result is less photo-realistic and not aligned with the desired segmap. FeNeRF struggles to provide such fine-grained editing. In terms of the impact region, it is clear that all models, except for \textsl{Pomo3D}, induce global alterations, notably IDE-3D and pix2pix3D in task(a) and particularly pix2pix3D in task(c), which causes a severe identity shift. By contrast, \textsl{Pomo3D} not only accomplishes all three tasks but also limits alternations only in the region of interest, benefiting from its object-aware synthesis scheme. 

\subsection{Ablation Study}

\begin{wrapfigure}{r}{0.35\textwidth}
\vspace{-5pt} 
\centering
\includegraphics[width=0.95\linewidth]{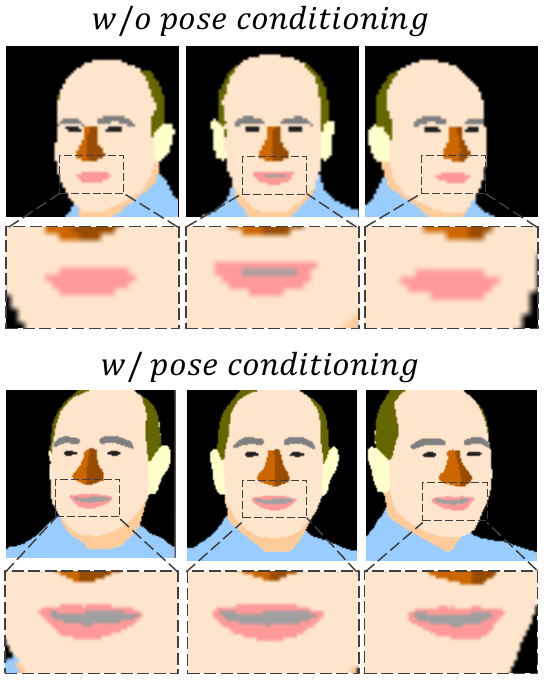} 
\caption{Pose-appearance association.}
\label{fig:F10}
\vspace{-30pt}  
\label{fig:wrapfig}
\end{wrapfigure}

\subsubsection{Pose Conditioning}
We first evaluate the effect of pose conditioning in the bias-conscious mapper. As illustrated in~\cref{fig:F10}, removing pose conditioning weakens the generator's sensitivity to camera poses, leading to changes in the geometry of the mouth as the viewpoint shifts. Instead, with pose conditioning, models can better decouple expressions and poses, enhancing geometry consistency.

\begin{figure}[tb]
  \centering
  \includegraphics[width=0.95\linewidth]{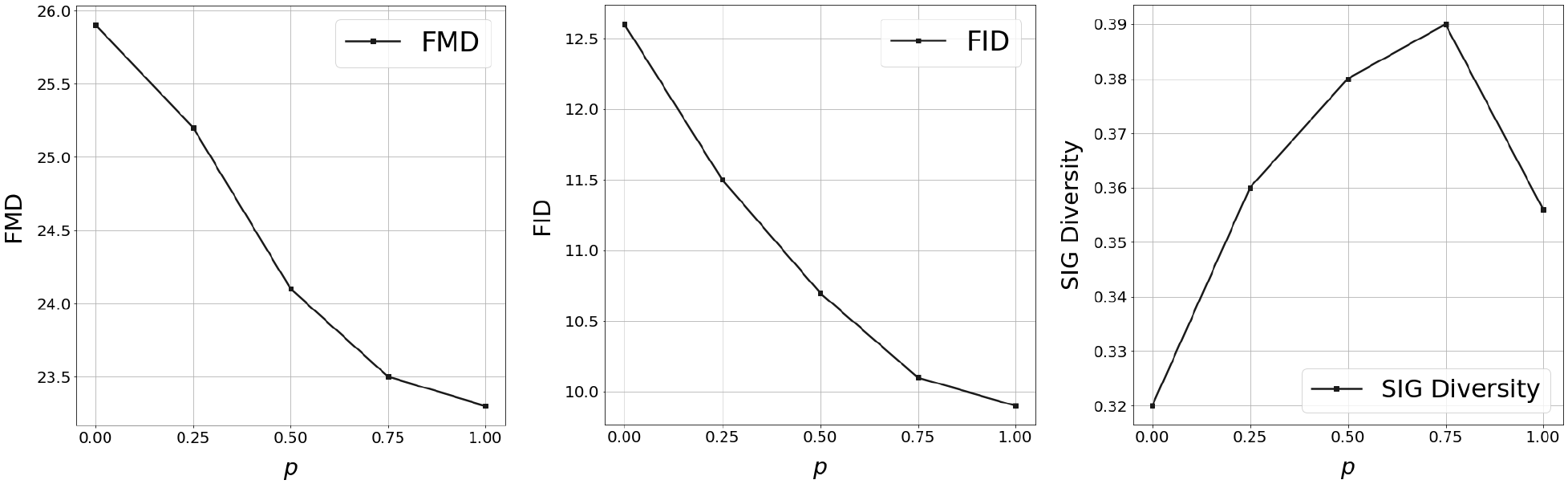}
  \caption{
  A study on the effects of identity conditioning. Here, \textit{p} is a training hyperparameter indicating the probability of sampling from $\mathcal{W}_{acc,g}$, while \textit{(1-p)} corresponds to the probability of sampling from $\mathcal{W}_{acc,g}^{*}$. Thus, \textit{p=0} and \textit{p>0} indicate the absence and presence of identity conditioning, respectively. The two figures on the left show the FMD and FID when training with various \textit{p} values, with lower values indicating that the generated results are closer to the training data distribution. The rightmost figure demonstrates the diversity of the produced accessory segmaps during inference.
  }
  \label{fig:F14}
\end{figure}

\subsubsection{Identity Conditioning}
\label{sec:Identity Conditioning}
In~\cref{fig:F14}, We further study the effect of identity conditioning in the bias-conscious mapper. During training, we introduce FMD (Fréchet Mask Inception Distance) to measure the distribution distance between real and generated accessory segmaps and FID for RGB images. During inference, SIG (Single Identity Generation) diversity is utilized to evaluate the diversity of accessory segmaps produced from a single identity by calculating the average LPIPS~\cite{zhang2018unreasonable} score. A larger hyperparameter \textit{p} indicates a greater extent of making the accessory generation identity-aware, enabling better capture of the biased associations between accessories and portraits during training, resulting in lower FMD and FID. However, always sampling from $\mathcal{W}_{acc,g}$ during training (i.e., \textit{p}=1) may lead to insufficient training of $\mathcal{W}_{acc,g}^{*}$. The embedding space of $\mathcal{W}_{acc,g}^{*}$ may grow arbitrarily and become less meaningful to the generator, ultimately leading to degenerate results when sampled during inference. Therefore, as mentioned in~\cref{sec:bias mapper}, we opt for \textit{p}=0.75 during training to achieve the best SIG diversity during inference. Experimental details and additional visual results are provided in the supplement.

\begin{wraptable}{r}{0.4\textwidth}
\vspace{0pt} 
\setlength{\abovecaptionskip}{-79pt} 
\centering
\resizebox{1.0\linewidth}{!}{
\begin{tabular}{@{}lccc@{}}
    \toprule[1pt]
            \midrule
                                 & FID$\downarrow$   & KID$\downarrow$   & FV-ID$\uparrow$ \\ \midrule
    \textbf{Baseline} & 10.4          & 2.78         & 0.705          \\
    \textbf{+S-Guidance} & 13.5          & 5.96         & 0.741          \\
    \textbf{+S-Encoder}  & \textbf{9.1}          & \textbf{1.51}         & \textbf{0.748}          \\ \bottomrule 

\end{tabular}}
\caption{Ablation study of the structure encoder.}
\label{tab:Estr}
\vspace{50pt}  
\end{wraptable}

\subsubsection{Structure Encoder}
In~\cref{tab:Estr}, we compare our full model with several variants. In the \textit{Baseline} scenario, texture generation is conditioned only on the combined segmap from the geometry model (namely, $\mathnormal{S}_{por}$ and $\mathnormal{S}_{acc}$) without the structure encoder. We modify the initial layers of the texture renderer to a seg2img architecture and leave the rest unchanged. Obviously, such generation only ensures semantic-level view consistency, thus yielding the lowest FV-ID. \textit{+S-Guidance} naively uses tri-plane-projected geometry features (combination of $\mathnormal{f}_{por}$ and $\mathnormal{f}_{acc}$) as the texture renderer's input with no structure encoder. We alter the input channel of the texture renderer to match that of $\mathnormal{f}_{por}$ and $\mathnormal{f}_{acc}$ while keeping the architecture unchanged. It leads to better FV-ID but introduces image quality degradation, indicating that the projected features primarily supervised for segmap generation are not a satisfactory structural prior. Instead, \textit{+S-Encoder} (full model) employs the structure encoder to transform geometry raw features into priors more suitable for texture generation, achieving better image quality while maintaining pixel-level view consistency.

\section{Conclusion}
This paper provides an effective solution to a less-addressed issue with \textsl{Pomo3D}, which offers the most extensive control over decorative objects of 3D portraits. It utilizes a bias-conscious mapper to create diverse accessories-portrait combinations and introduces a method to transform user's scribbles into 3D accessories. The training dataset used in this method is publicly available for future studies. \textsl{Pomo3D} has various applications, such as 3D accessory virtual try-ons, opening new possibilities for further research. More experimental details, applications, limitations, and discussions can be found in the supplementary material.

\bibliographystyle{splncs04}


\newpage

\section{Overview}
\label{sec:overview}

We present additional results as a supplement to the main paper. Firstly, in~\cref{imp}, we present further details of the main paper's experiments and PAC-Mask. Secondly, additional experiments and comparisons are provided in~\cref{add exp}. Thirdly, we introduce applications in~\cref{app}. Then we discuss the model's limitations and ethical considerations in~\cref{lim,eth}. Lastly, more visual results are displayed in~\cref{add result}.

\section{Implementation Details}
\label{imp}

\subsubsection{Dual Geometry Tri-planes}
The portrait geometry tri-plane $\mathnormal{f}_{por}^{triplane}$ is largely based on EG3D~\cite{chan2022efficient} and utilizes the same hyperparameters as stated in the original paper. Each branch of the feature adapter $\varphi_{i}$ consists of two blocks, each containing two modulated convolution layers, modulated by the accessory geometry code $\mathnormal{w}_{acc,g}$. The per-pixel classifier $\mathnormal{F}_{toSEG}$ is implemented by two multi-layer perceptrons (MLPs) with no output activation and about 6k trainable parameters. 

\subsubsection{Structure Encoder}
Within the structure encoder $\mathnormal{E}_{str}$, portrait features $\mathnormal{f}_{por}$ and accessory features $\mathnormal{f}_{acc}$ are first combined into a single feature map based on the binary mask $\mathnormal{m}_{acc}$. Namely, $\mathnormal{f}_{combined} = \mathnormal{m}_{acc} \odot \mathnormal{f}_{acc} + \left( 1 - \mathnormal{m}_{acc} \right) \odot \mathnormal{f}_{por}$. Then, we use two residual blocks to learn to form the structural prior $\mathnormal{f}_{fused}$ from combined features $\mathnormal{f}_{combined}$. Besides, since not all samples are accessorized, to fit the training data distribution better, we adaptively set the probability of $(Accs == True)$ based on the ratio of accessories in datasets during training (for instance, about 0.37 for CelebAMask). When $(Accs == False)$, $\mathnormal{m}_{acc}$ is an all-zeros mask; otherwise, $\mathnormal{m}_{acc}$ is a binary mask indicating the accessory region.

\subsubsection{Texture Renderer}
In the texture renderer $\mathnormal{R}_{tex}$, features in each block are modulated by two styles and then re-combined based on binary masks $\mathnormal{m}_{acc}$ and $(1 - \mathnormal{m}_{acc})$, then followed by a spatially-adaptive normalization. To further edit textures other than accessories, we switch among three schemes of compositional synthesis during training. When $(Accs == True)$, $\mathnormal{m}_{acc}$ is set as the binary mask of accessories. When $(Accs == False)$, we set $\mathnormal{m}_{acc}$ either as an all-zeros mask (same as the structure encoder) or as a binary mask for other decorative objects, such as clothing, hair, etc. (different from the structure encoder). In other words, even without adding accessories, we randomly render two texture styles on different semantic regions of the portrait, thereby enhancing the disentanglement of these semantic regions. These decorative objects share the same texture style with accessories, i.e., $\mathnormal{w}_{acc,t}$. As such, the accessory texture code $\mathnormal{w}_{acc,t}$ is extended to encompass the texture style for other decorative objects, not just accessories.

\subsubsection{Multiple Accessory Wearing}
In cases where $\mathcal{N}$ accessories are worn simultaneously, we forward the accessory branch (i.e., $\mathnormal{G}_{acc}$) $\mathcal{N}$ times with $\mathcal{N}$ different accessory geometry codes $\mathnormal{w}_{acc,g}$, obtaining $\mathcal{N}$ sets of $(\mathnormal{f}_{acc}$, $\mathnormal{S}_{acc},\mathnormal{m}_{acc})$ while keeping the same portrait geometry (i.e., the same $\mathnormal{w}_{por,g}$). Let $\mathnormal{m}_{acc}^{U}$ be the union of these $\mathcal{N}$ accessory binary mask, $\mathnormal{m}_{acc}^{U} = \mathnormal{m}_{acc}^{0} \cup \mathnormal{m}_{acc}^{1} \cup \mathnormal{m}_{acc}^{2} \cdots \cup \mathnormal{m}_{acc}^{\mathcal{N}}$. Inside the structure encoder, the combined feature map $\mathnormal{f}_{combined}$ is now composed of a single $\mathnormal{f}_{por}$ and $\mathcal{N}$ pairs of $(\mathnormal{f}_{acc}$, $\mathnormal{m}_{acc})$.  

\begin{equation}
\mathnormal{f}_{combined} = \left( 1 - \mathnormal{m}_{acc}^{U} \right) \odot \mathnormal{f}_{por} + \sum_{n=0}^{\mathcal{N}} \mathnormal{m}_{acc}^{n} \odot \mathnormal{f}_{acc}^{n}
\end{equation}

Here, $\mathnormal{f}_{acc}^{n}$ denotes the projected accessory features generated by the $\mathnormal{n}^{th}$ accessory geometry code. In the texture renderer, the original $\mathnormal{m}_{acc}$ is replaced with $\mathnormal{m}_{acc}^{U}$, with all else remaining unchanged.

\subsubsection{Bias-Conscious Mapper}
In order to make the accessory generation identity-aware, we introduce a cross-attention module. The original accessory code $\mathnormal{w}_{acc,g}^{*}$ serves as the query, while the identity embedding acts as both key and value. Let $\mathnormal{W}_{A}$ and $\mathnormal{W}_{I}$ denote the original code and the identity embedding, respectively. $\mathnormal{W}_{Q}$, $\mathnormal{W}_{K}$ and $\mathnormal{W}_{V}$ are projection matrices. The cross-attention process can be formalized as follows: 
$\mathnormal{softmax} (( \mathnormal{W}_{Q}\mathnormal{W}_{A}) (\mathnormal{W}_{K}\mathnormal{W}_{I})^{T})
\mathnormal{W}_{V}\mathnormal{W}_{I} $

\begin{figure}[t]
  \centering
  \includegraphics[width=\linewidth]{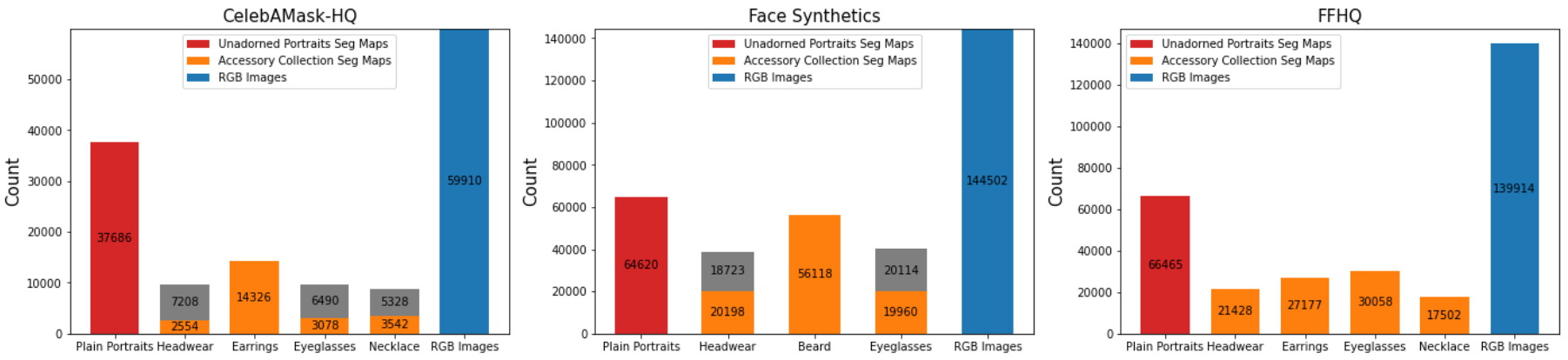}
  \caption{
  Statistics for PAC-Mask. The gray bars represent the augmented data.
  }
  \label{fig:F25_B}
\end{figure}

\subsubsection{Dataset Details: PAC-Mask}
\begin{wrapfigure}{r}{0.5\textwidth}
\vspace{-20pt} 
\centering
\includegraphics[width=\linewidth]{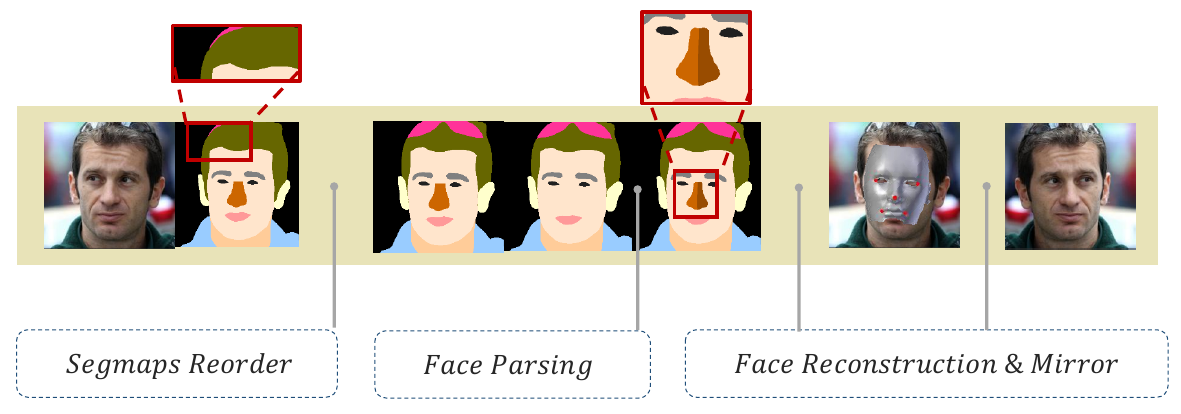} 
\caption{Data Pre-processing. (left to right)}
\label{fig:F25_A}
\vspace{-20pt}  
\label{fig:wrapfig}
\end{wrapfigure}

We curate PAC-Mask from the widely accessible dataset, CelebAMask-HQ~\cite{lee2020maskgan}, FFHQ~\cite{karras2019style} and FaceSynthetics~\cite{wood2021fake}. The main pre-processing involves three steps as in~\cref{fig:F25_A}. First, we notice that many existing mask-based portrait synthesis methods directly train with single-channel semantic maps, obtained by stacking binary masks of each semantic region (e.g., hair, brow, lip) from the raw annotations (such as CelebAMask-HQ) along the channel axis and returning the indices of the maximum values of each pixel along the axis. This process makes each pixel represent one semantic type, compressing multiple regional semantic masks into a single semantic map, and facilitating easier data processing. Yet, there may be some potential issues. In areas of overlap, a single pixel may contain more than one semantic type. This leads to the overlap area being dominated by a certain semantic (based on the order of channel index) and the loss of information from other semantics. As illustrated in the first step of~\cref{fig:F25_A}, if glasses are placed over hair, this processing may cause the glasses to disappear due to the overlap with the hair. Therefore, we reorder the semantics, and when other semantics overlap with accessory semantics, we prioritize the accessories to prevent them from being overlooked. Second, we split the nose semantics into two parts, the left and right nose, for clearer geometry with an off-the-shelf network~\cite{yu2018bisenet}, considering the geometry tri-plane is primarily learned from a large quantity of single-view semantic maps. Third, we obtain the pose of each image through face reconstruction and mirror all images for data augmentation.

\begin{figure}[t]
  \centering
  \includegraphics[width=\linewidth]{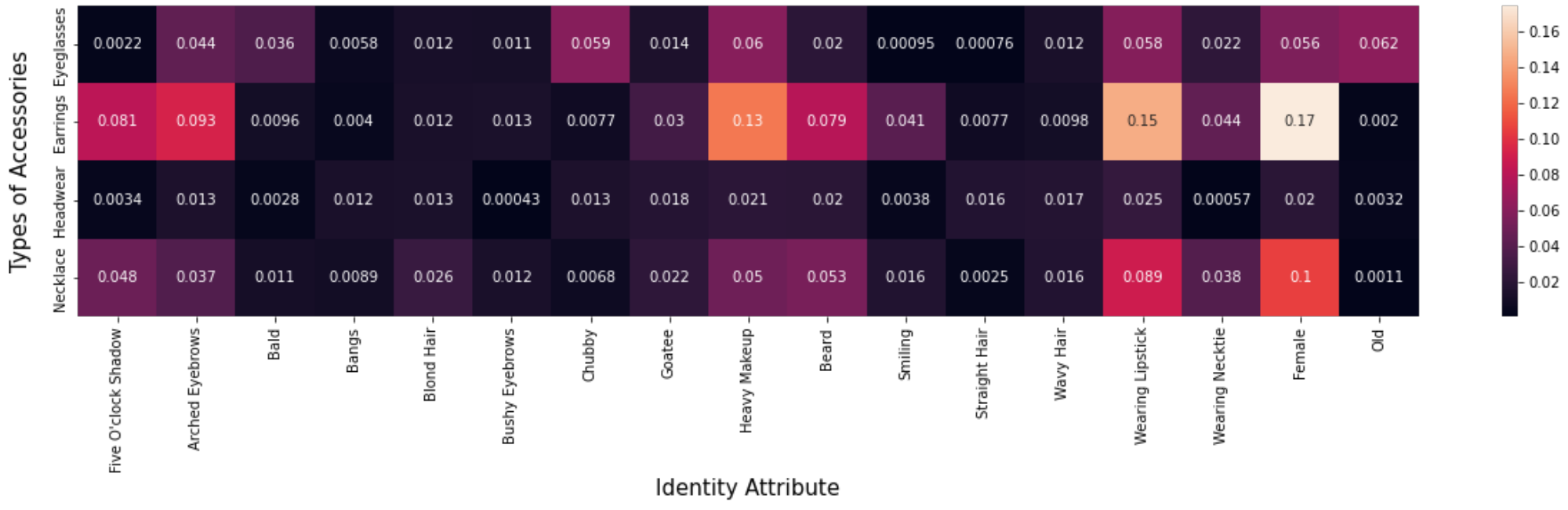}
  \caption{
  A heatmap of the mutual information between accessory types and corresponding portrait attributes. The minimum value is zero, with higher values indicating stronger associations.
  }
  \label{fig:F25_C}
\end{figure}

Next, we categorize all semantic maps into two groups based on the presence of accessories. In the case of semantic maps containing accessories, only the accessory portion is retained, while the remaining parts (face, skin, etc.) are discarded. Consequently, this results in the data being partitioned into three non-overlapping groups: accessory collection segmaps (segmentation maps), unadorned portrait segmaps, and RGB images. We list the quantities of three datasets after processing in~\cref{fig:F25_B}. Due to severe imbalances in the types of accessories in CelebAMask-HQ and FaceSynthetics, we either randomly extract one accessory from the semantic maps of multiple accessories to form a separate semantic map or randomly generated new duplicate samples to mitigate the data imbalance, represented by gray bars in the figure. In FaceSynthetics, we also categorize beards as accessories and use the same generation pipeline.

Moreover, we investigate the correlation between accessories and portrait attributes within PAC-Mask. We record the type of each accessory and its corresponding portrait attributes to calculate the mutual information, which measures the similarity between two categorical variables. As illustrated in~\cref{fig:F25_C}, we can clearly observe certain biased associations between specific accessories and portrait attributes. The most notable are the associations of earrings and necklaces with gender-related attributes, such as Earrings-Female, Earrings-WearingLipsticks, Earrings-HeavyMakeup, and Necklace-Female associations. \\This further validates the necessity of using a bias-conscious mapper.

\begin{figure}[t]
  \centering
  \includegraphics[width=0.7\linewidth]{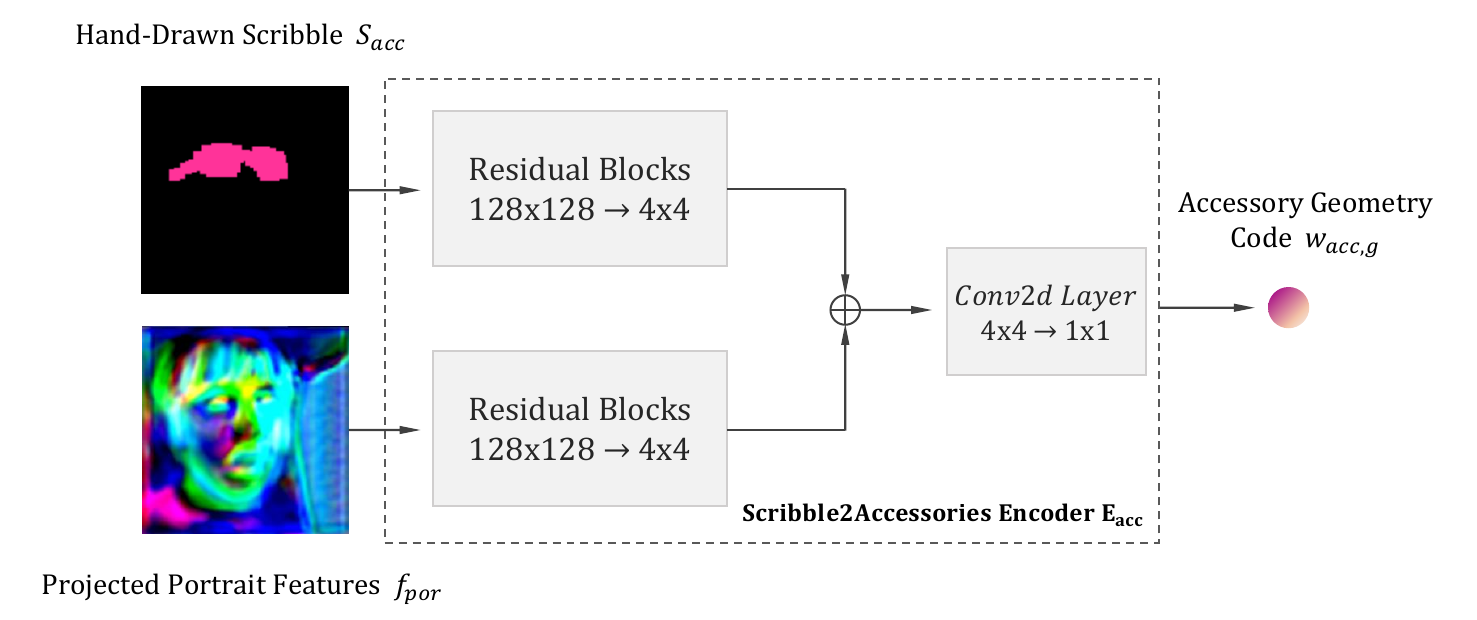}
  \caption{
  The architecture of the Scribble2Accessories encoder. $\oplus$ denotes element-wise addition.
  }
  \label{fig:F26}
\end{figure}

\subsubsection{Scribble2Accessories}
We demonstrate the architecture of the Scribble2Access-ories encoder in~\cref{fig:F26}. The encoder predicts the accessory geometry code $\mathnormal{w}_{acc,g}$ based on the hand-drawn scribble map $\mathnormal{S}_{acc}$ and the projected portrait features $\mathnormal{f}_{por}$ as the pose and structural information of the wearer. The dimensions of the accessory geometry code $\mathnormal{w}_{acc,g}$ and the code length in the accessory codebook $\mathcal{A}$ are both 256. 

\subsubsection{Training Details}
To ensure a stable and fast training convergence, we initially pre-train the segmap part (geometry) and then train the whole model. Following~\cite{karras2020analyzing}, a non-saturating logistic loss~\cite{goodfellow2020generative} with R1 regularization~\cite{mescheder2018training} is applied. The batch size is set to 16 for $512^{2}$ training, and the learning rate is $2.5e-3$. Most of our training parameters are inherited from StyleGAN2. It takes about three days to get visually pleasing results when trained on 8 Tesla V100 GPUs.

\subsubsection{Experimental Details}
In this section, we provide more implementation details of the experiments from the main paper, including the quantitative evaluations in the \textit{Comparison} section and the identity conditioning in the \textit{Ablation Study}. 

In the quantitative evaluations, we compare our \textsl{Pomo3D} with other SOTA mask-based 3D portrait editing models in terms of image quality, RGB-segmap alignment, and view consistency. For quality evaluation (FID and KID), we randomly sample 2k segmaps and generate ten different textured RGB images for each segmap. For RGB-segmap alignment, we randomly generate 1k pairs of segmaps and RGB images for computing mIoU and Acc. The semantic maps of RGB images are provided by an off-the-shelf network~\cite{yu2018bisenet}. As for view consistency, for each model, 100 random facial identities with 12 pre-defined camera poses are used to calculate FV-ID.

In the ablation study of identity conditioning, we use FMD (Fréchet Mask Inception Distance) to measure the distribution distance between real and generated segmaps and SIG (Single Identity Generation) diversity to evaluate the diversity of generated accessory segmaps. In practice, for FMD, we generate 10k accessory segmaps to calculate the Fréchet distance with real accessory segmaps, using the early layers of a pre-trained mask discriminator as inception layers. For SIG diversity, We first randomly sample 1k portrait identities (i.e., 1k $\mathnormal{w}_{por,g}$). Each will be sampled with two kinds of accessories (i.e., two different $\mathnormal{w}_{acc,g}$) to form a segmap pair for LPIPS evaluation.

\begin{figure}[t]
  \centering
  \includegraphics[width=\linewidth]{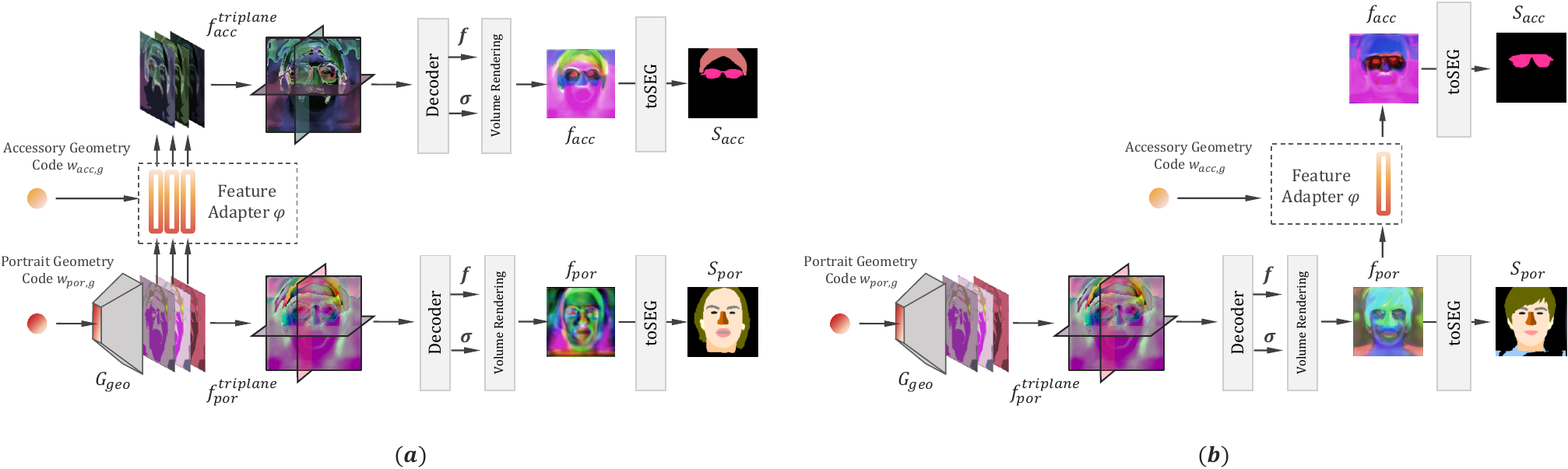}
  \caption{
  Visual results of the feature adapter. As illustrated in (b), the feature adapter intensifies the focus on certain areas based on different accessory codes.
  }
  \label{fig:F24}
\end{figure}

\section{Additional Experiments and Visualization}
\label{add exp}
\begin{wrapfigure}{r}{0.33\textwidth}
\vspace{-20pt} 
\centering
\includegraphics[width=\linewidth]{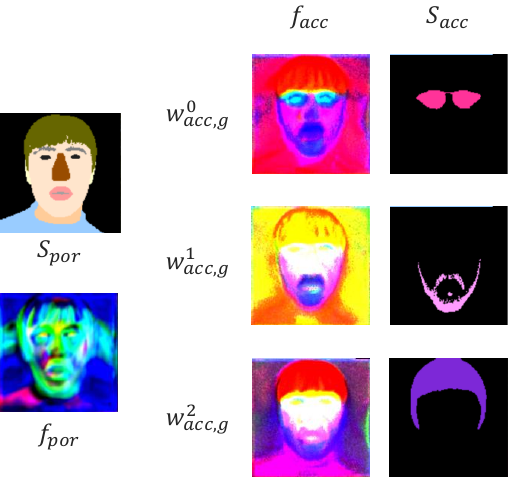} 
\caption{Given certain $\mathnormal{S}_{por}$ and $\mathnormal{f}_{por}$, we observe $\mathnormal{f}_{acc}$ and $\mathnormal{S}_{acc}$ with different $\mathnormal{w}_{acc,g}$.}
\label{fig:F34}
\vspace{-20pt}  
\label{fig:wrapfig}
\end{wrapfigure}

\subsubsection{Feature Adapter}
To better visualize the function of the feature adapter, as illustrated in~\cref{fig:F24}, we modify the architecture from (a) to (b). We employ one of the feature adapter's branches, keep the architecture unchanged, and re-train it. The feature adapter's three branches originally map each plane's features in the portrait geometry tri-plane $\mathnormal{f}_{por}^{triplane}$ to the accessory geometry tri-plane $\mathnormal{f}_{acc}^{triplane}$. Now, the single-branch feature adapter directly maps the projected portrait features $\mathnormal{f}_{por}$ to the accessory space on the image plane. As such, we can observe the influence of the feature adapter on the portrait features $\mathnormal{f}_{por}$.

As shown in~\cref{fig:F34}, the feature adapter shifts the spatial focus based on different accessory geometry codes $\mathnormal{w}_{acc,g}$. Thus, distinct accessory semantic maps $\mathnormal{S}_{acc}$ can be generated by the following per-pixel classifier $\mathnormal{F}_{toSEG}$. Such an accessory generation scheme can also yield results similar to those in ~\cref{fig:F24}(a). 
Nevertheless, it disregards the projective geometry, resulting in accessories that only function in 2D. When the viewpoint changes, visible misalignments between the portrait and generated accessories can be easily observed. Thus, as shown in~\cref{fig:F24}(a), rather than operating on the 2D plane, we employ the feature adapter across the tri-plane to shift the focus among 3D volumes, ensuring stronger view consistency.

\begin{figure}[t]
  \centering
  \includegraphics[width=0.8\linewidth]{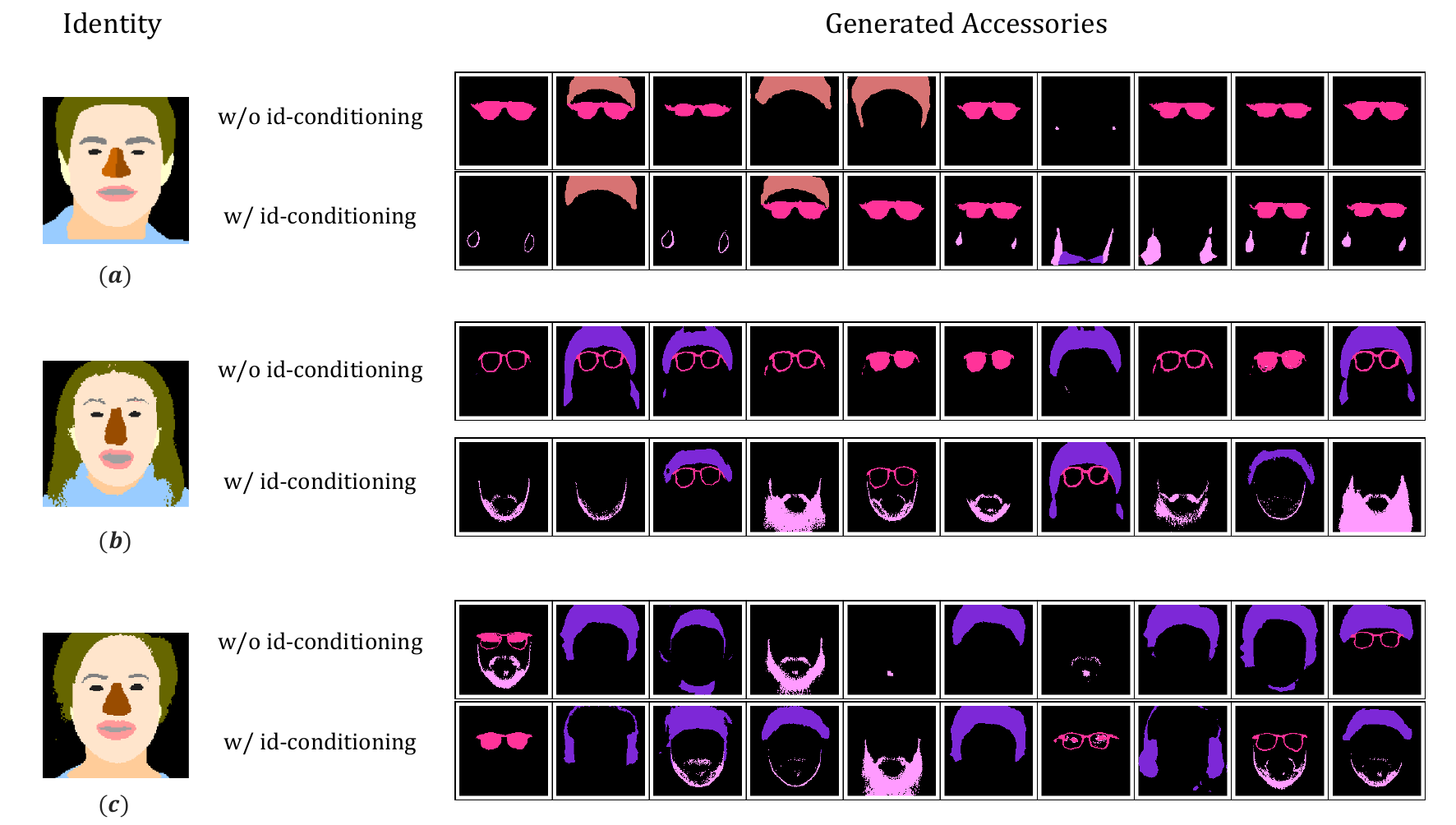}
  \caption{
  The visual result of identity conditioning. With identity conditioning, we can create more diverse combinations of accessories and wearers.
  }
  \label{fig:F28}
\end{figure}

\subsubsection{Identity Conditioning}

We present the visualization results of identity conditioning in~\cref{fig:F28}. Next to each identity, there are two rows of accessory samples produced by the top ten seeds during inference. The top row shows results without identity conditioning, while the bottom row indicates results with identity conditioning. It can be observed that without identity conditioning, the variety of accessories is usually limited and often related to the external attribute of the identity. For example, for a more masculine appearance (a), earrings are rarely produced in the top row; or for a more feminine appearance (b), there are no beard-related items in the top row. Conversely, with identity conditioning, as shown in the bottom row, a wider variety of accessories can be obtained. Therefore, identity conditioning greatly enhances the possibility of creating new combinations of accessories and wearers even beyond the representation of the dataset.

\subsubsection{Ablation Study of SPADE}
\begin{wrapfigure}{r}{0.5\textwidth}
\vspace{-30pt} 
\centering
\includegraphics[width=\linewidth]{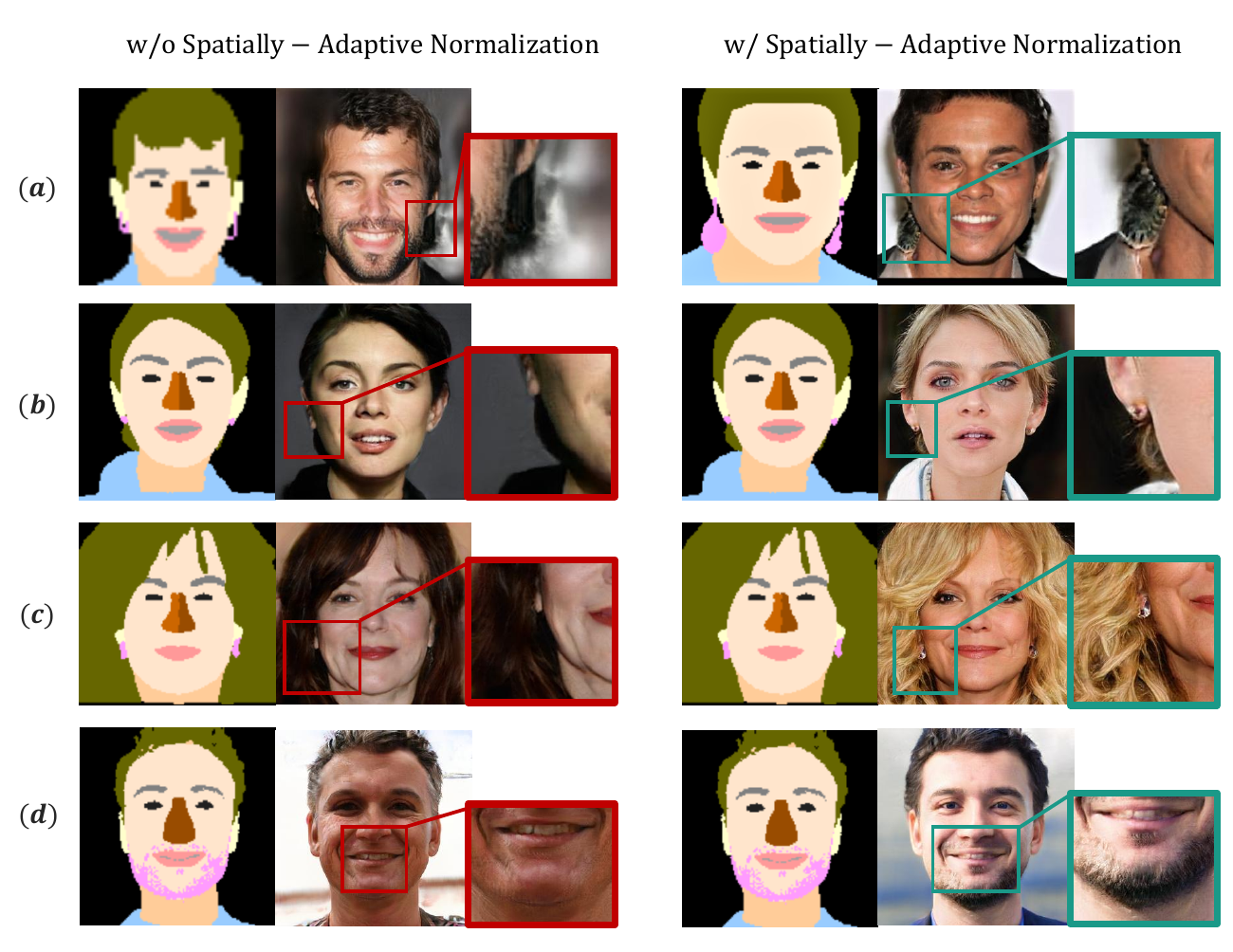} 
\caption{Visualization of SPADE}
\label{fig:F29}
\vspace{-70pt}  
\label{fig:wrapfig}
\end{wrapfigure}
We notice that even with the generated semantics of accessories, relatively small accessories may still be overlooked during texture generation due to the holistic nature of the global discriminator. Therefore, we further introduce spatially-adaptive normalization (SPADE)~\cite{park2019semantic} into our texture rendering block to impose semantic-aware constraints on generated features, ensuring that the RGB images match the desired segmentation maps.

As seen in ~\cref{fig:F29}(b) and (c) without SPADE, accessories, particularly small ones, may be overlooked during texture generation. Moreover, in (a) without SPADE, even if we generate out-of-distribution segmap combinations (a masculine portrait with earrings), it may still be challenging to produce the corresponding texture. In contrast, with SPADE, we can align the generated results with the desired segmap, including earrings in (a), (b), and (c) and beards in (d). Nevertheless, as seen in (a) with SPADE, for samples that go beyond what the training dataset can represent, the texture may still exhibit noticeable artifacts, which remains another issue to be resolved.

\begin{figure}[t]
  \centering
  \includegraphics[width=0.9\linewidth]{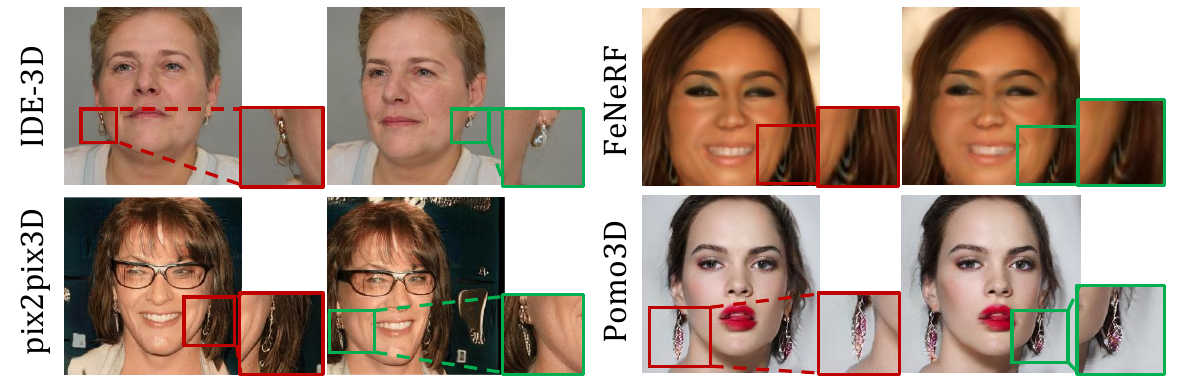}
  \caption{
  Comparison of the visual quality of earrings.
  }
  \label{fig:F15}
\end{figure}

\subsubsection{Comparison of Earrings}
We compare the visual quality of earrings generated by four methods in~\cref{fig:F15}. IDE-3D\cite{sun2022ide} is limited to producing small and inconspicuous earrings near the earlobe. FENeRF\cite{sun2022fenerf}'s results more closely resemble parts of hair rather than earrings. pix2pix3D\cite{deng20233d} tends to create more fragmented patterns. On the other hand, \textsl{Pomo3D} is capable of generating prominent and complete earrings.

\section{Applications} 
\label{app}
\subsubsection{3D Accessory Virtual Try-on}

\begin{figure}[t]
  \centering
  \includegraphics[width=\linewidth]{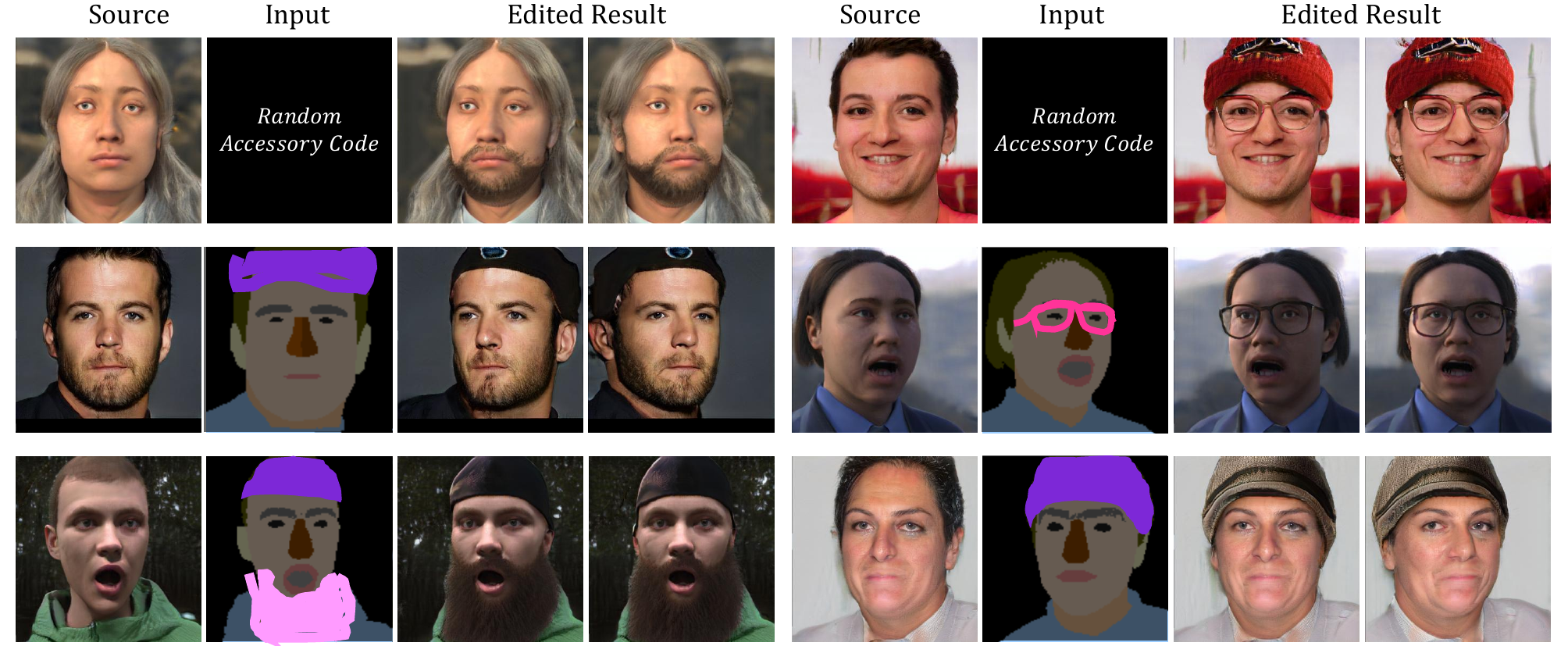}
  \caption{
  Accessories and beards can be generated either by a random accessory geometry code (first row) or by the user's scribble map (bottom two rows). They can be created from any viewpoint, not limited to the frontal view.
  }
  \label{fig:F22}
\end{figure}

Virtual try-on technology allows consumers to virtually try on clothes or accessories, finding out how these products appear on them without any physical interaction~\cite{islam2024deep,cheng2021fashion}. Through our dedicated accessory branch, we extend 3D portrait synthesis to broader applications such as 3D portrait accessorizing or 3D accessory virtual try-ons, opening new possibilities for further research. As shown in~\cref{fig:F22}, \textsl{Pomo3D} is capable of producing diverse accessories on specified portraits. With Scribble2Accessories, users can first draft a rough design of the accessory, and then choose a preferred texture.

\subsubsection{Interactive Avatar Customization}
\begin{figure}[t]
  \centering
  \includegraphics[width=\linewidth]{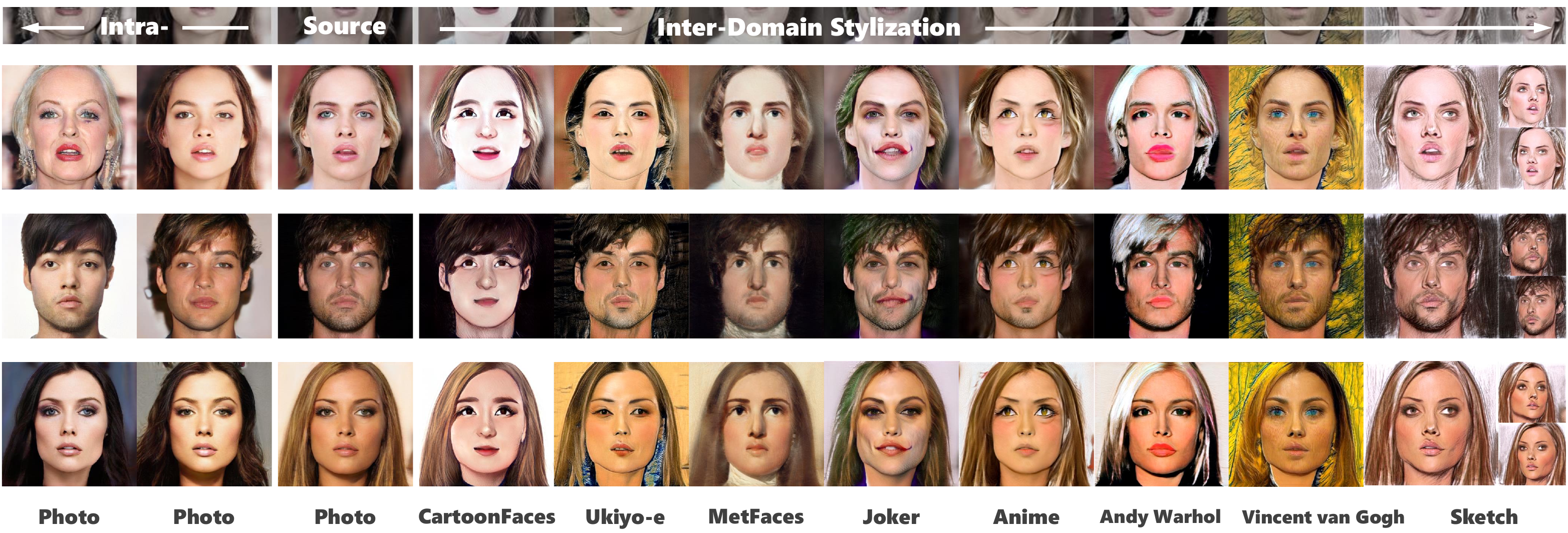}
  \caption{
  Inter- and intra-domain stylization. \cite{CartoonFaces, karras2020training}
  }
  \label{fig:F16}
\end{figure}
\begin{figure}[t]
  \centering
  \includegraphics[width=\linewidth]{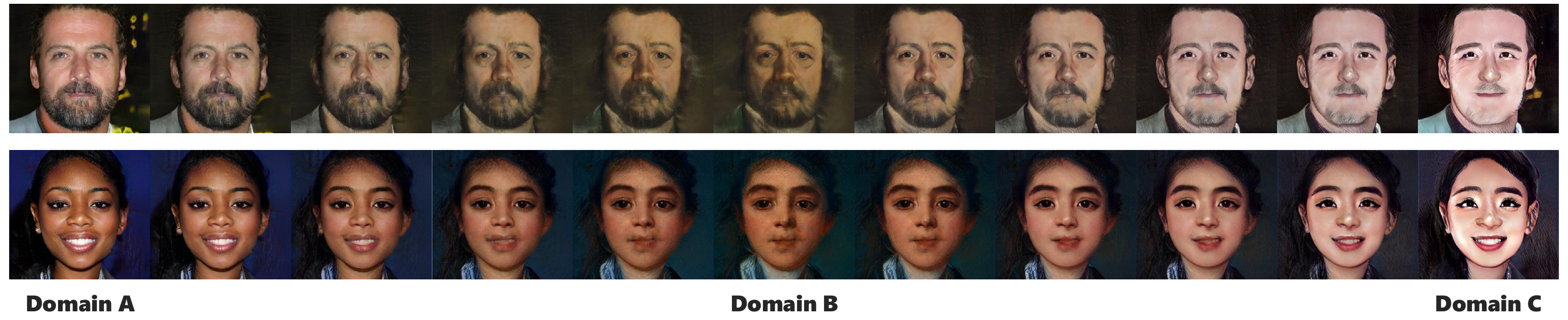}
  \caption{
  Smooth avatar stylization through model weights interpolation.
  }
  \label{fig:F17}
\end{figure}

\begin{figure}[t]
  \centering
  \includegraphics[width=\linewidth]{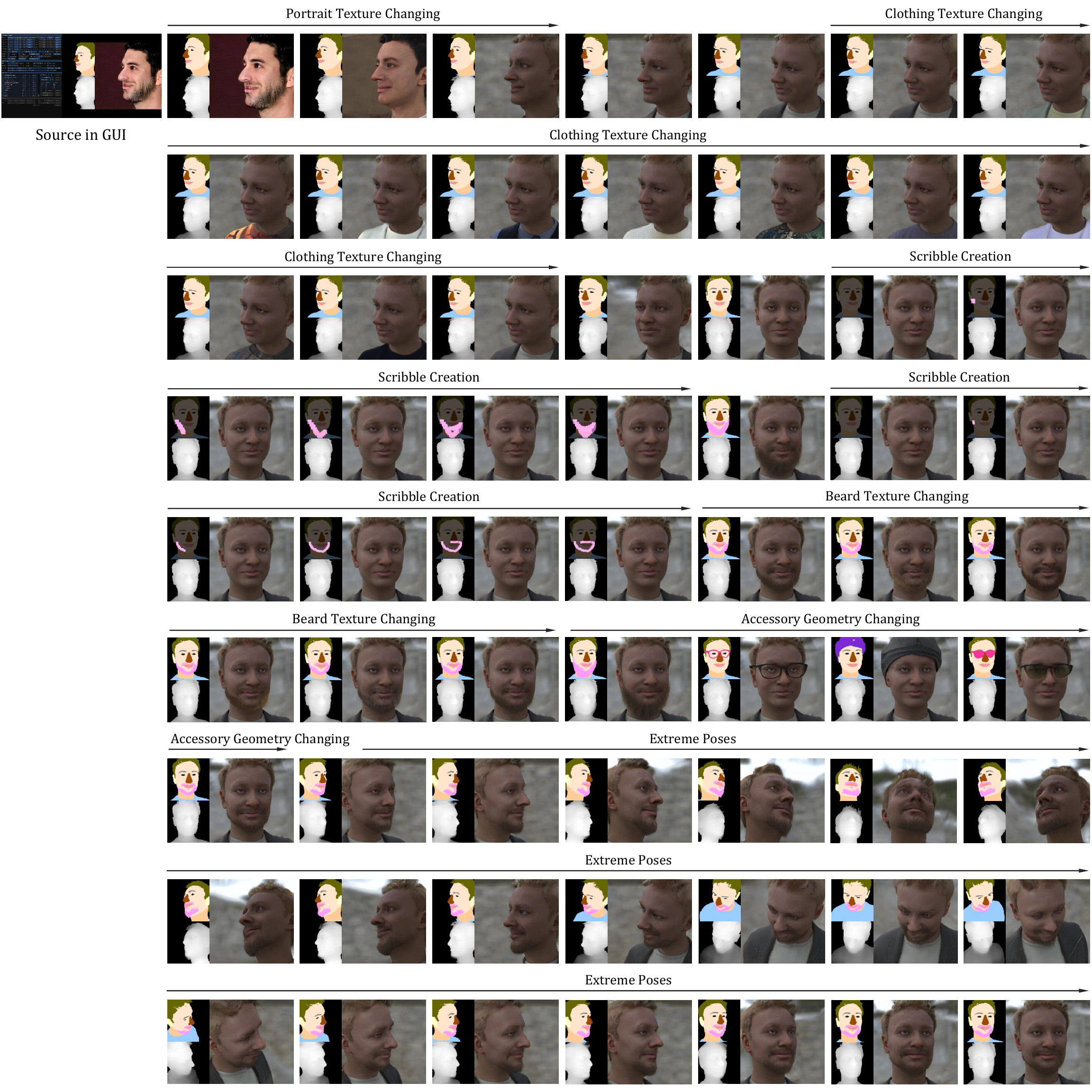}
  \caption{
  Examples of sequential edits within the GUI interface. To better display the results, we remove the control panel on the left. Users can perform various edits directly in the provided GUI, including scribble creation. The editing results will be 3D consistent, even under extreme poses.
  }
  \label{fig:F21_B}
\end{figure}

We provide a GUI interface for real-time portrait generation featuring comprehensive controllability, integrating explicit camera control, global and local adjustments of geometry or texture, inter- or intra-domain stylization, and Scribble2Accessories at an interactive framerate, resulting in a distinct and instant creation of stylized avatar in a virtual world. The GUI interface is based on~\cite{gu2021stylenerf,karras2021alias}, and we extend it to offer a broader range of editing functionalities. In~\cref{fig:F21_B}, we demonstrate an example of sequential editing through the GUI interface. 

To further achieve avatar stylization, we employ two approaches for domain adaptation while preserving the capability to detach accessories. The first approach is to fine-tune the texture renderer with limited novel domain data ($\sim$100 images) while fixing the remaining part, thanks to the model's disentanglement of geometry and texture generation. If the number of domain images is insufficient, the differentiable augmentation technique \cite{zhao2020differentiable} is adopted to prevent overfitting. The second approach follows \cite{gal2021stylegan} to adapt the pre-trained generator to textually-prescribed domains, which leverages the semantic power of large-scale Contrastive-Language-Image-Pretraining (CLIP) models \cite{radford2021learning}. As shown in~\cref{fig:F16,fig:F17}, we provide the functionality to switch between different styles in the GUI interface.

Since the FaceSynthetics dataset has a wider pose distribution and semantic annotations for beards, we can employ the two approaches mentioned above to adapt models trained on FaceSynthetics (synthetic domain) back into the real image domain with the CelebAMask-HQ or FFHQ datasets, achieving beard generation and improved results for large poses in the real image domain. The source image shown in~\cref{fig:F21_B} is an example of a real-image-domain portrait in an extreme pose (profile).

\section{Limitations}
\label{lim}
While \textsl{Pomo3D} is capable of producing more diverse accessories, there are still some potential issues that need to be addressed. For instance, the quality of certain accessories, particularly necklaces with their significant variation and limited training data, is not entirely satisfactory. Moreover, \textsl{Pomo3D} also struggles to generate complex patterns on headwear. With larger and more varied training datasets of accessories, we believe these problems can be greatly alleviated. Additionally, this generation pipeline makes the inversion of the entire image more difficult. In practice, we first invert the segmap and fix its geometry code, then search for the optimal texture code. A more effective solution may be to train a separate encoder that maps entire images back to the generator’s latent space. We leave it for future work.

\section{Ethical Concerns}
\label{eth}
\textsl{Pomo3D} possesses the capability to generate and manipulate 3D portraits. Therefore, there are risks of misuse, such as identity fraud or causing misrecognition in facial recognition systems. This capability should be used carefully and must be regulated. There are several ongoing research on deepfakes detection~\cite{masood2023deepfakes,mubarak2023survey}, which aim to distinguish between synthesized faces and real faces. The training data we used and our generated results can also aid in training for deepfakes detection.

\section{Additional Results}
\label{add result}
\begin{figure}[t]
  \centering
  \includegraphics[width=\linewidth]{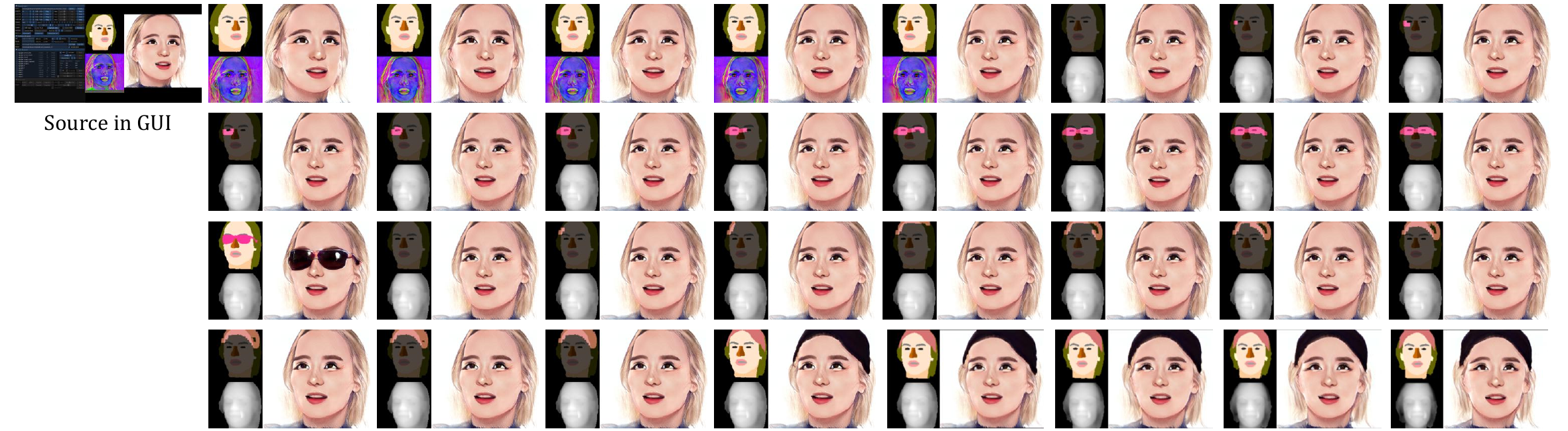}
  \caption{
  Additional sequential edits on stylized portraits.
  }
  \label{fig:F21_A}
\end{figure}

\begin{figure}[t]
  \centering
  \includegraphics[width=\linewidth]{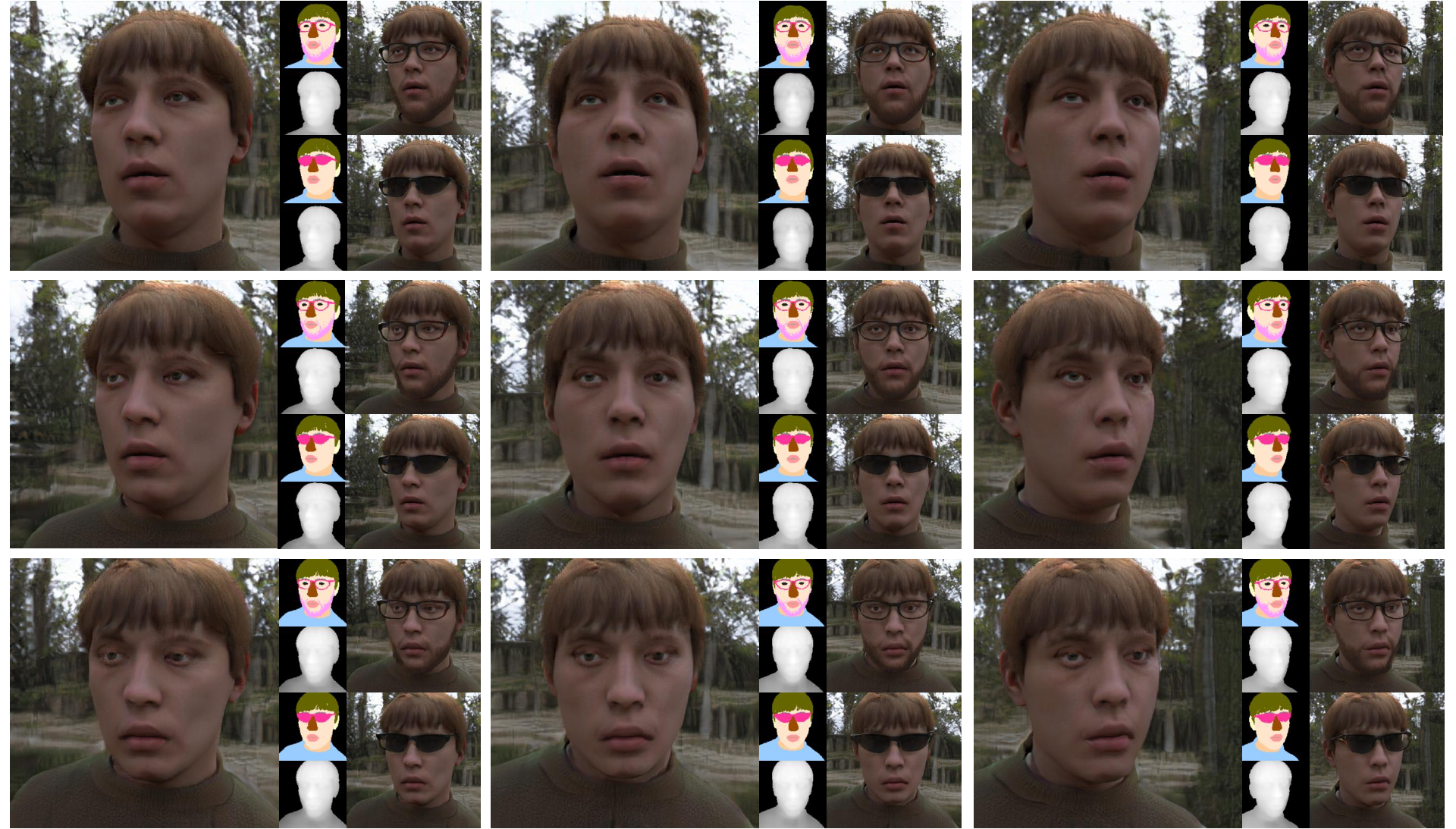}
  \caption{
  In each cell of the nine grid, we display the source image along with two cases of accessory-wearing results. The first includes glasses and a beard, and the second includes sunglasses.
  }
  \label{fig:F32}
\end{figure}

\begin{figure}[t]
  \centering
  \includegraphics[width=\linewidth]{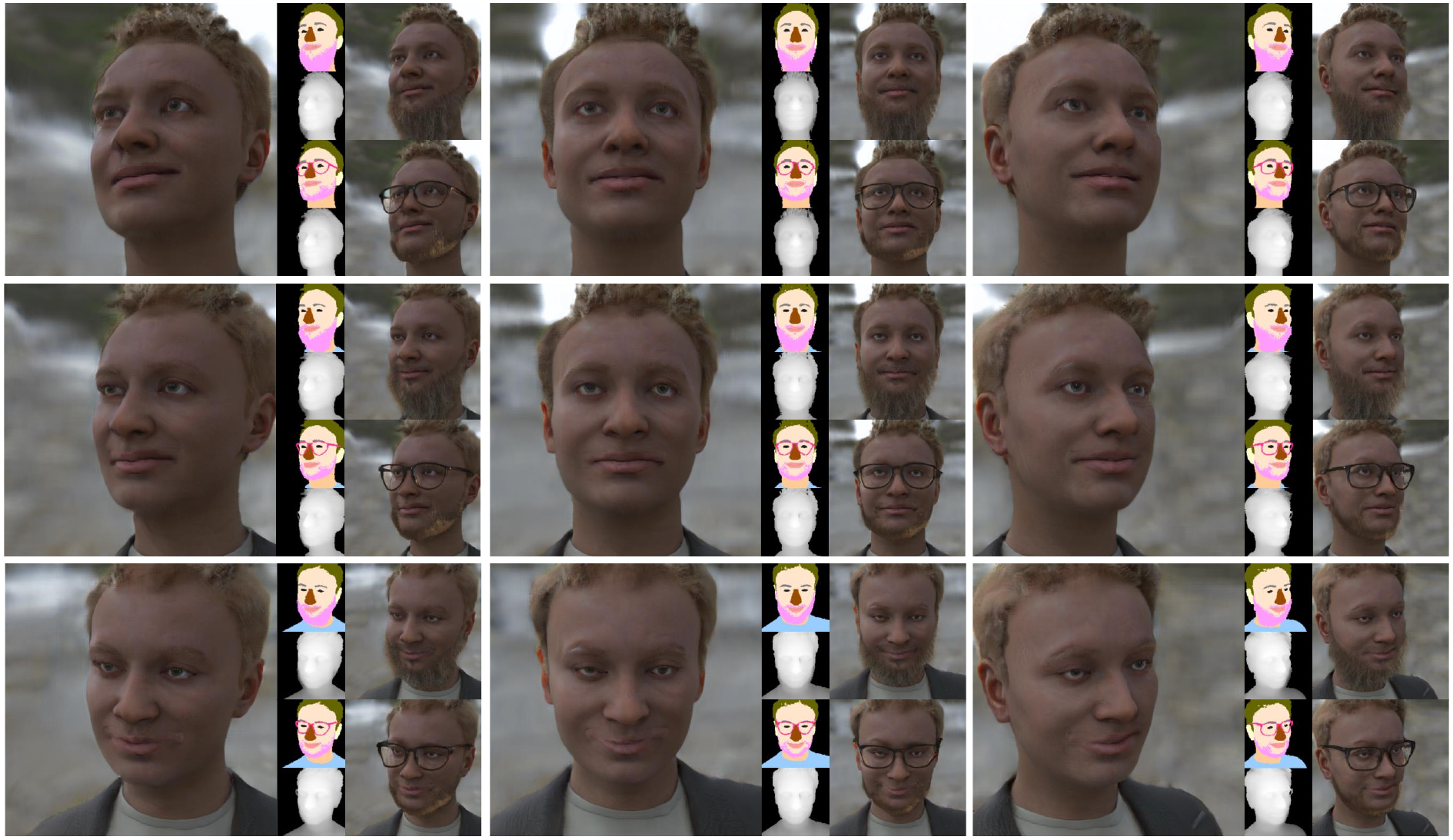}
  \caption{
  In each cell of the nine grid, we display the source image along with two cases of accessory-wearing results. The first includes a full beard, and the second includes glasses and a beard with a different texture.
  }
  \label{fig:F18}
\end{figure}

\begin{figure}[t]
  \centering
  \includegraphics[width=\linewidth]{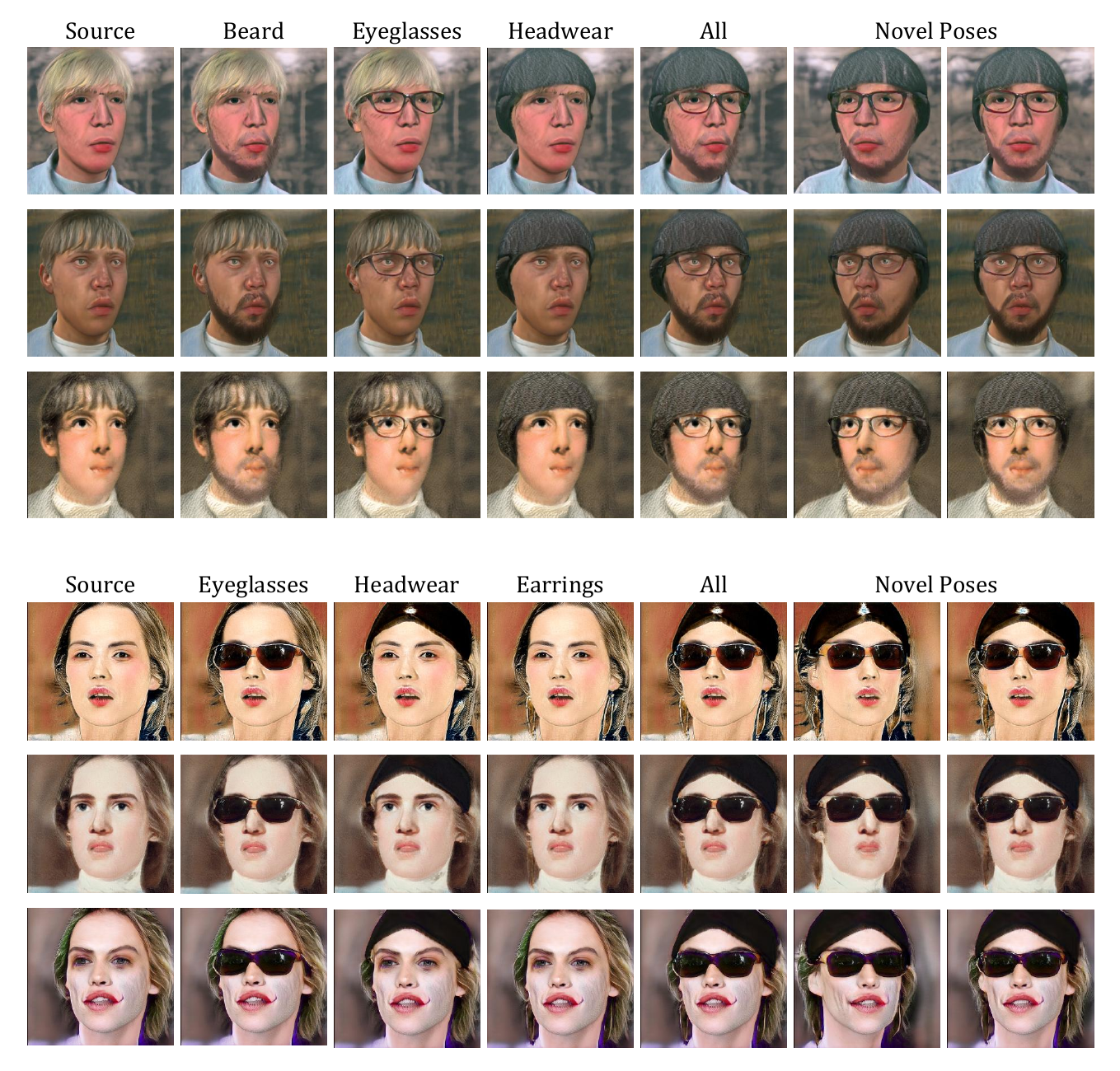}
  \caption{
  Accessory wearing on stylized portraits.
  }
  \label{fig:F31}
\end{figure}

\begin{figure}[t]
  \centering
  \includegraphics[width=\linewidth]{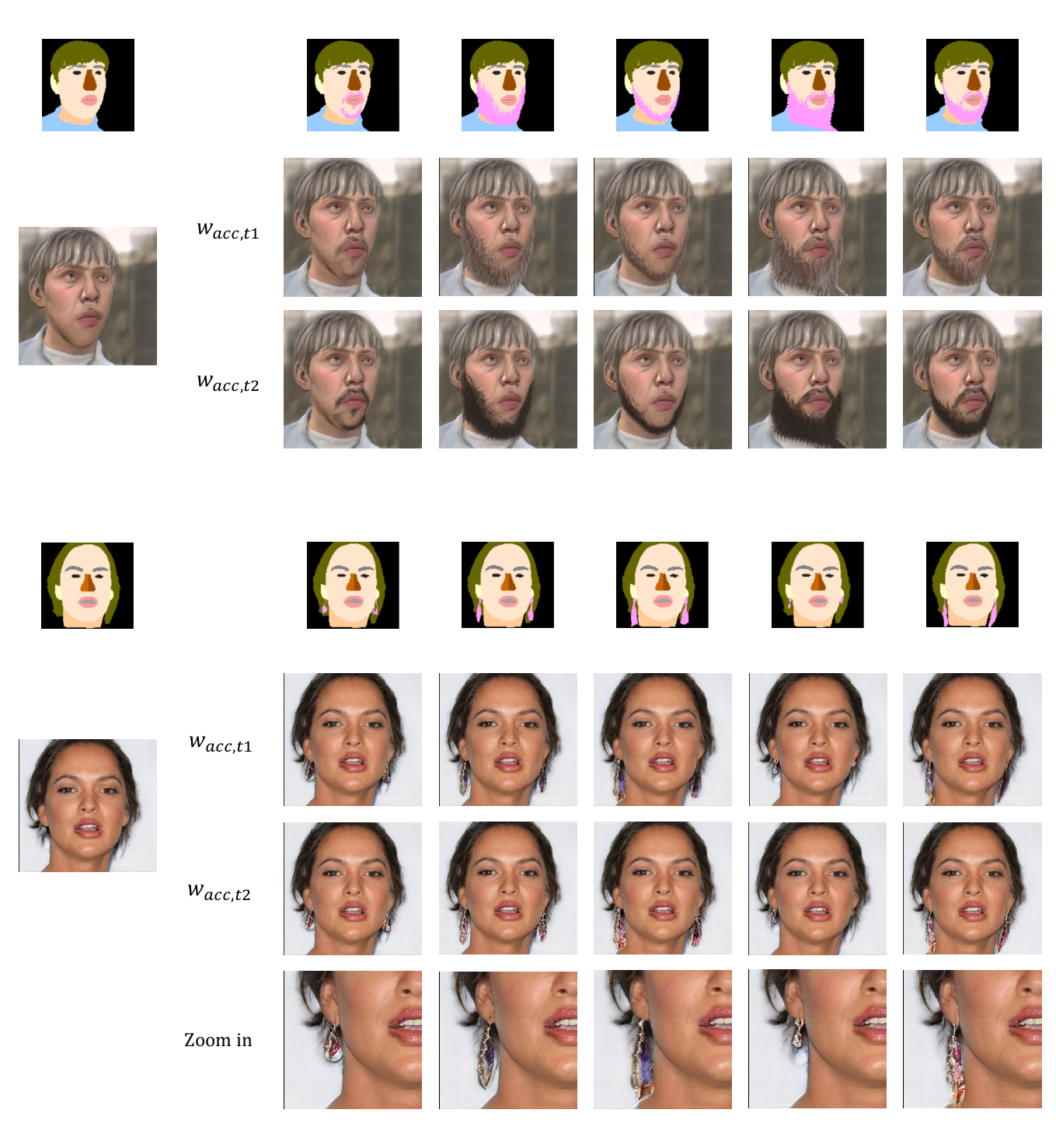}
  \caption{
  Diverse shapes and textures of accessories and beards. On the left are the source images, and on the right, we show the source images with various shapes of beards and earrings, paired with two different texture codes $\mathnormal{w}_{acc,t1}$ and $\mathnormal{w}_{acc,t2}$. This allows users to create virtual avatars with unique and personalized accessories.
  }
  \label{fig:F33}
\end{figure}

\begin{figure}[t]
  \centering
  \includegraphics[width=\linewidth]{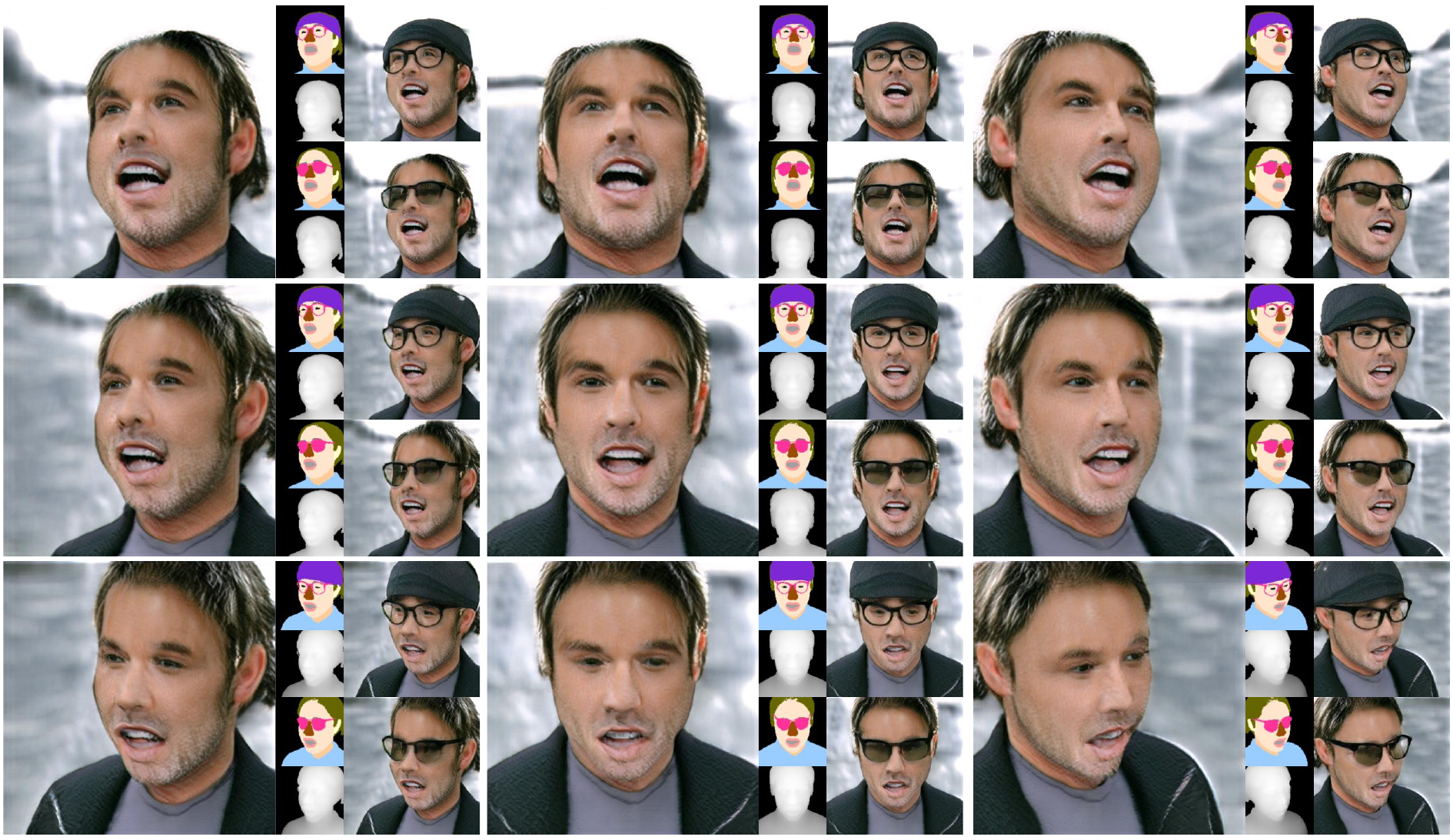}
  \caption{
  In each cell of the nine grid, we display the source image along with two cases of accessory-wearing results. The first includes headwear and glasses, and the second includes glasses with different lens textures. 
  }
  \label{fig:F19}
\end{figure}

\begin{figure}[t]
  \centering
  \includegraphics[width=\linewidth]{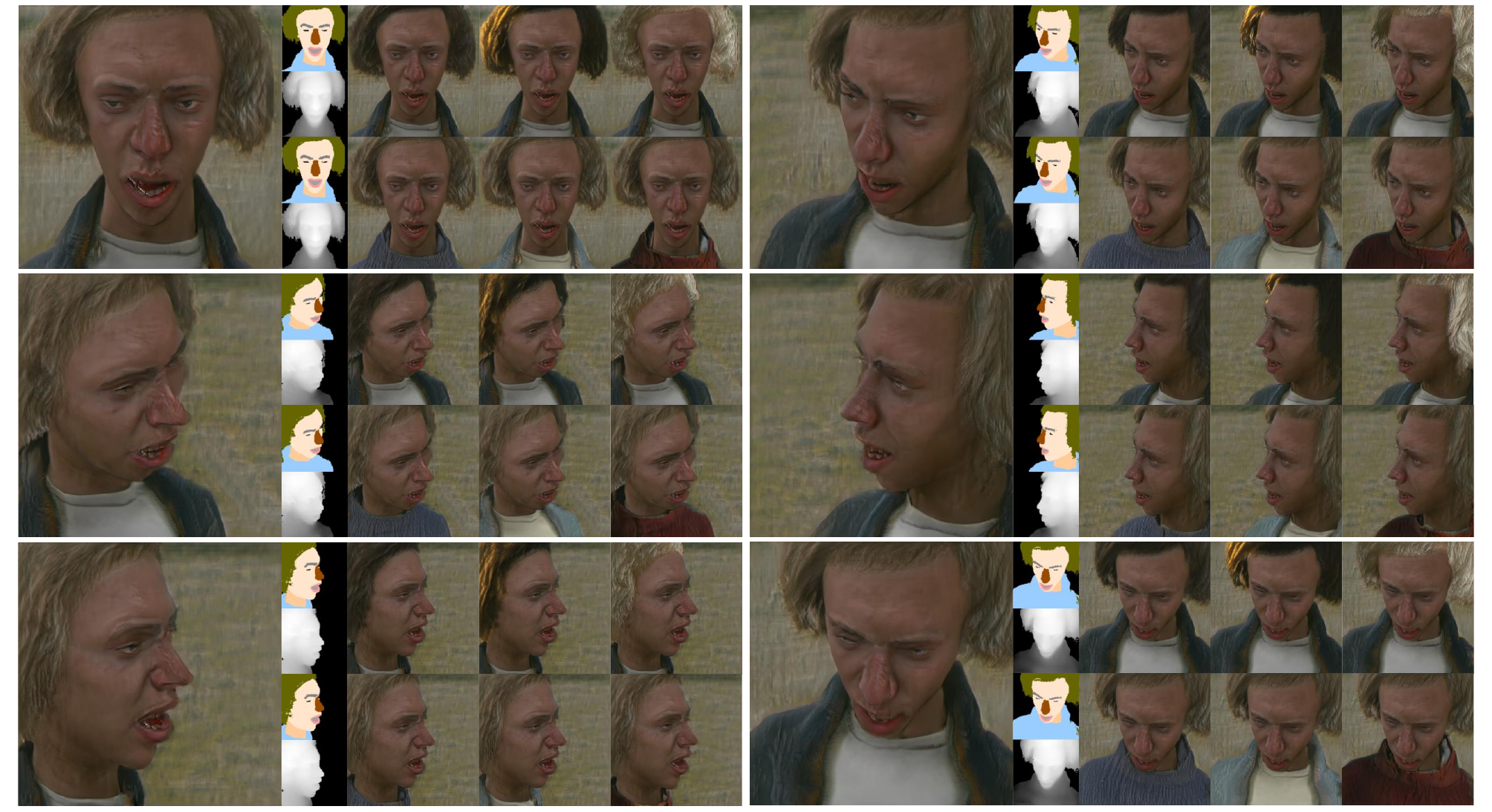}
  \caption{
  View-consistent texture editing. We demonstrate the results of the same portrait under six different camera poses. In each pose, we change the texture of the hair (top row) and clothes (bottom row).
  The editing results are consistent across different poses.
  }
  \label{fig:F20}
\end{figure}

\begin{figure}[t]
  \centering
  \includegraphics[width=0.7\linewidth]{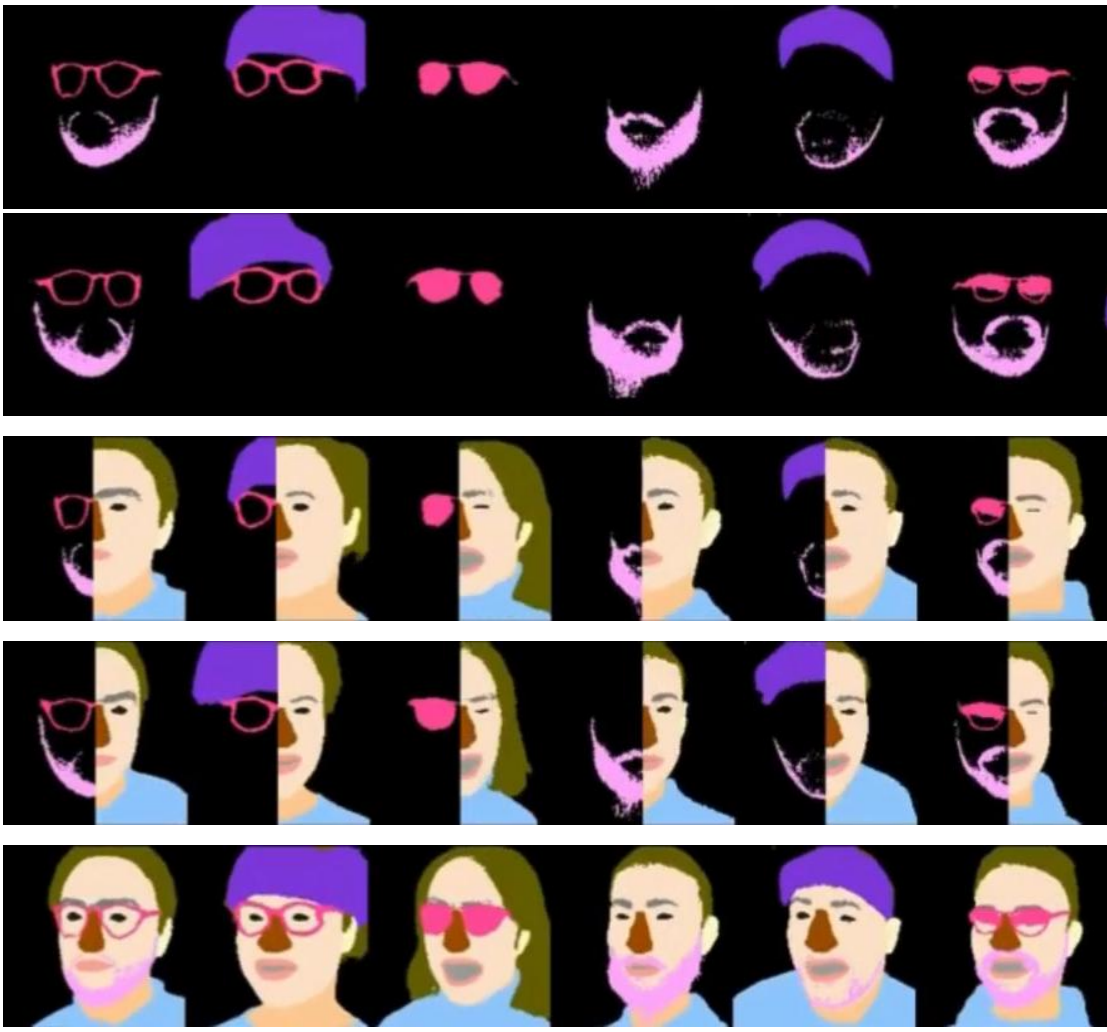}
  \caption{
  We combine the 3D accessory segmap with the 3D portrait segmap to achieve a 3D consistent accessory-wearing portrait.
  }
  \label{fig:F27}
\end{figure}

In this section, we provide additional visual results as a supplement to the main paper. We demonstrate additional sequential edits on stylized portraits (\cref{fig:F21_A}), accessory wearing in a nine-grid layout from different viewpoints (\cref{fig:F32,fig:F18,fig:F19}), accessory wearing on stylized portraits (\cref{fig:F31}), diverse shapes and textures of accessories (\cref{fig:F33}), view-consistent texture editing (\cref{fig:F20}), and a demonstration of accessories implicitly aligned with portraits (\cref{fig:F27}).

\end{document}